\documentclass[acmsmall]{acmart}

\setcopyright{cc}
\setcctype{by}
\acmJournal{PACMMOD}
\acmYear{2026} \acmVolume{4} \acmNumber{6 (SIGMOD)} \acmArticle{317}
\acmMonth{12} \acmDOI{10.1145/3856312}

\AtBeginDocument{%
  }

\usepackage{algorithm}
\usepackage{algpseudocode}
\usepackage{amsmath}
\usepackage{cleveref}
\usepackage{tabularx}
\usepackage{subcaption}
\usepackage{xcolor}
\usepackage{xspace}

\newcommand{\sys}[1]{{\Sys}}
\newtheorem{observation}{Observation}
\newcommand{\Sys}{\textsc{MapFlow}\xspace}

\definecolor{revisionmeta}{HTML}{000000}
\definecolor{revisionone}{HTML}{000000}
\definecolor{revisiontwo}{HTML}{000000}
\definecolor{revisionthree}{HTML}{000000}

\newcommand{\revieweronechange}[2][]{#2}
\newcommand{\reviewertwochange}[2][]{#2}
\newcommand{\reviewerthreechange}[2][]{#2}

\begin{document}

\title{Accelerating Cross-Model LLM Inference via Attention Map Reuse}

\author{Zhan Jin}\authornotemark[1]
\email{zhanjin2004@gmail.com}
\affiliation{%
  \institution{Peking University}
  \city{Beijing}
  \country{China}
}

\author{Yuze Luo}\authornotemark[1]
\email{luoyuze@stu.pku.edu.cn}
\affiliation{%
  \institution{Peking University}
  \city{Beijing}
  \country{China}
}

\author{Bin Cui}\authornote{School of Computer Science \& Beijing Key Laboratory of Software and Hardware Cooperative Artificial Intelligence Systems, Peking University}\authornote{Key Lab of High Confidence Software Technologies (MOE), Peking University}
\email{bin.cui@pku.edu.cn}
\affiliation{%
  \institution{Peking University}
  \city{Beijing}
  \country{China}
}

\author{Xupeng Miao}\authornotemark[1]
\email{xupeng.miao@pku.edu.cn}
\affiliation{%
  \institution{Peking University}
  \city{Beijing}
  \country{China}
}

\renewcommand{\shortauthors}{Jin et al.}

\begin{abstract}
LLM-augmented data workflows often involve multiple heterogeneous models operating over substantial shared context, such as schemas, tool specifications, interaction history, and intermediate results. Although prefix caching can reduce prefilling overhead, existing approaches are largely limited to reuse within the same model because KV states are tightly coupled to model-specific architectures.
We present \Sys, a system for efficient heterogeneous LLM serving via cross-model attention map reuse. Instead of reusing KV states, \Sys reuses attention maps, which capture token dependency patterns and are more transferable across models. To enable efficient and accurate reuse, we combine partial recomputation with lightweight linear translation across layers and heads. To make attention reuse practical for long contexts, we further introduce block-level sparse compression and a fused retrieval kernel.
Experiments show that \Sys reduces prefilling latency and improves end-to-end serving efficiency while preserving generation quality across heterogeneous model pairs and multi-turn routing workloads. These results suggest that attention maps provide an effective reusable state for heterogeneous LLM serving, especially in prefix-heavy data workflows. Our code is available: \url{https://github.com/PKU-DAIR/MapFlow}.

\end{abstract}

\ccsdesc[500]{Information systems~Database management system engines}
\ccsdesc[500]{Information systems~Information storage systems}
\ccsdesc[500]{Computing methodologies~Artificial intelligence}

\keywords{Large language models; Heterogeneous LLM serving; Cross-model prefix reuse; Attention map reuse; LLM-augmented data systems; Multi-LLM workflows}

\received{17 April 2026}
\received[revised]{19 August 2026}
\received[accepted]{12 September 2026}

\maketitle

\section{Introduction}

Large language models (LLMs) are increasingly becoming a core building block of modern data systems. Recent studies~\cite{hu2023chatdb,weng2025bridgescope,giannakouris2024demonstrating,deng2025alayadb,zhang2025plforge,dorbani2025beyond,DBLP:journals/dase/WangCWULG25,DBLP:journals/dase/ZhuZSSY25,DBLP:conf/icde/ZhangLMWCLDC26} have explored using LLMs for database interaction and symbolic memory, query optimization and tuning, data preprocessing, and entity matching, suggesting that LLMs are no longer merely front-end interfaces but active components in data management pipelines. In these settings, LLM execution is rarely a single prompt-response step. Instead, requests are embedded in multi-stage workflows that repeatedly expose models to long, structured contexts, including database schemas, system constraints, tool specifications, prior interaction history, and intermediate results. As these contexts accumulate across turns and operators, repeatedly processing the same prefix becomes a major bottleneck for interactive latency and system throughput~\cite{DBLP:journals/csur/MiaoOZCJCJ26,hu2023chatdb,weng2025bridgescope,miao2024spotserve,shi2026multi}.

When multiple requests share the same prompt prefix and are served by the same model, cached Key-Value (KV) tensors can be reused to avoid redundant prefilling and reduce Time-to-First-Token (TTFT), as demonstrated by recent LLM serving systems such as vLLM~\cite{Kwon2023vllm} and SGLang~
\cite{lian2024Sglang}.
However, this assumption does not hold in many LLM-augmented data systems, where a single data workflow often involves \textbf{multiple heterogeneous LLMs} or agents.
To balance quality, cost, and response time, these systems assign different subtasks to different models, such as using a stronger model for planning or semantic interpretation and smaller models for refinement, formatting, or summarization.
Such heterogeneous execution appears naturally in routing-based assistants, cascaded inference, and multi-agent workflows~\cite{Ong2025routellm,jiang2023llmblender,wadlom2026efficient}.
Although these executions frequently share substantial prompt prefixes, existing \textbf{prefix caching} mechanisms are largely restricted to \textit{identical model instances}. Once execution moves across models, even identical database-centric context must be recomputed from scratch, leading to high latency and poor reuse.

This problem is particularly acute in data management workloads. Compared with open-ended chat, many database-facing prompts are highly structured and persist across multiple requests. A text-to-SQL or database assistant may repeatedly prepend the same schema and system instructions; a tuning assistant may repeatedly condition on workload summaries and engine constraints; and an agentic analytics pipeline may pass along tool traces, metadata, and prior reasoning results~\cite{llmopt2025}. These workloads therefore create a distinctive combination of prefix-heavy execution and cross-model transitions. From a data systems perspective, this is not merely a serving inefficiency; it is a missed opportunity for intermediate state reuse across operators in heterogeneous LLM pipelines.

A straightforward idea is to reuse cached KV states across models. Prior work on cross-model or multi-model execution has explored routing, blending, multiplexing, and state reuse, but KV tensors remain tightly coupled to model-specific hidden representations, dimensions, structures, and parameters~\cite{liu2024droidspeak,yao2024cacheblend,fu2025cachetocache,dery2026latent,zhu2024muxserve,sheng2024slora,miao2024specinfer}.
This coupling makes them difficult to translate across heterogeneous architectures without incurring either substantial accuracy loss or nontrivial adaptation overhead. As a result, existing approaches struggle to achieve the accuracy-efficiency trade-off required by online systems. This limitation suggests a broader question: \textit{what form of intermediate state can be reused across heterogeneous LLMs while remaining efficient enough for real-time execution?}

In this paper, we answer this question with \Sys, an efficient heterogeneous LLM serving system based on cross-model attention map reuse. Our key insight is that, \textit{while KV states are model-specific latent representations, attention maps encode token-to-token dependency structure in Transformer models}. For models that share the same tokenizer and process the same context, these dependency patterns can exhibit stronger \textbf{structural similarity} across models than KV tensors do. This makes attention maps a more suitable reusable state for heterogeneous execution. Based on this observation, \Sys transfers attention maps from a sender model to a receiver model, uses lightweight linear translation to bridge differences in layers and heads, and combines this with partial recomputation to preserve semantic fidelity.

Turning this idea into a practical runtime raises two systems challenges. First, raw attention maps have quadratic size in context length, making naive caching prohibitively expensive.
Second, even if cached, directly reconstructing dense receiver-side attention would introduce excessive memory traffic and runtime overhead.
\Sys addresses these challenges through a co-designed system runtime.
It applies a hardware-friendly block-level top-$p$ sparse compression scheme to store attention maps in a compact block-sparse layout, and uses a fused retrieval kernel that combines attention translation and value aggregation without materializing dense intermediate tensors.
The system further organizes cached sparse blocks with prefix-hash-based indexing and overlaps retrieval with partial recomputation, enabling efficient online reuse within modern serving stacks.

We evaluate our system across heterogeneous model pairs and multi-turn workloads. Our results show that \Sys reduces TTFT by 15\%–20\% while maintaining generation quality, and improves end-to-end throughput by up to 25\% under the same latency target. We further show that its benefits are especially pronounced in long-context, multi-turn settings, where repeated workflow-level context amplifies the cost of standard prefilling. These results suggest that cross-model attention reuse can serve as an effective execution primitive for LLM-augmented data systems.

Our paper makes four main contributions:
\begin{itemize}
    \item We identify cross-model prefix reuse as a fundamental bottleneck in LLM-augmented data workflows, where long structured contexts are repeatedly processed across heterogeneous models.
    \item We propose attention maps as a reusable intermediate execution state for heterogeneous LLM serving, avoiding the tight architectural coupling of KV-cache-based reuse.
    \item We design \Sys, a practical LLM serving system that combines partial recomputation, lightweight attention-map translation, sparse block-level compression, and fused online retrieval.
    \item We demonstrate through extensive experiments that \Sys improves latency and throughput for heterogeneous multi-model execution, making it a promising systems substrate for prefix-heavy LLM data workflows.
\end{itemize}

\section{Background}

\label{sec:background}

\begin{figure}[h]
    \centering
    \includegraphics[width=0.6\linewidth]{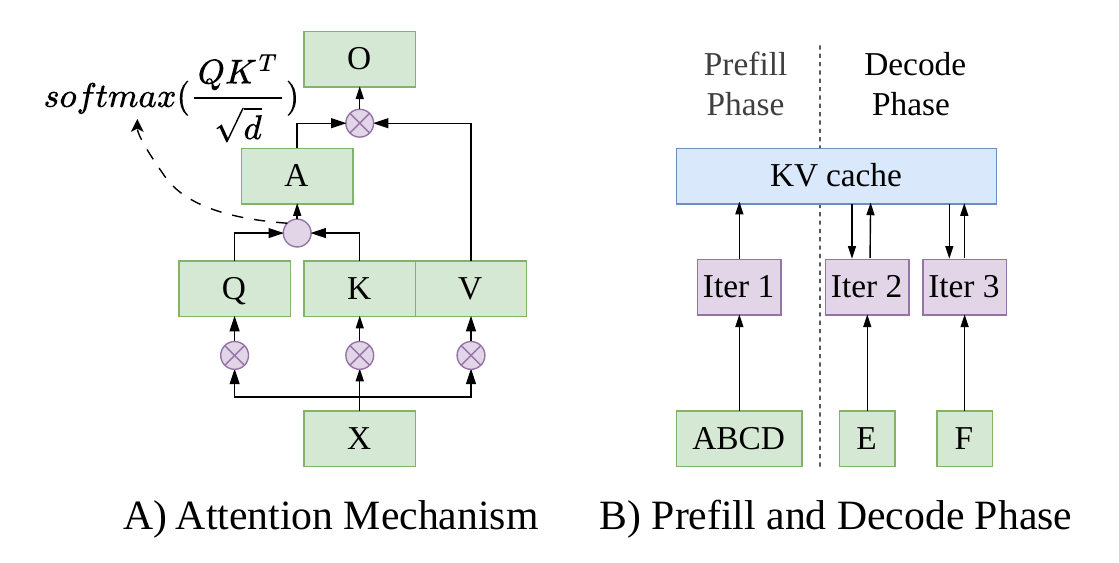}
    \caption{Illustration of attention mechanism, prefill phase and decode phase.}
    \label{fig:background}
\end{figure}

\subsection{Transformer and Multi-Head Attention}

Modern Large Language Models (LLMs) are predominantly built upon the Transformer architecture \cite{vaswani2017attention}, utilizing a decoder-only configuration. A standard Transformer block consists of two primary components: Multi-Head Attention (MHA) and a Feed-Forward Network (FFN).
As shown in Figure~\ref{fig:background}A, the MHA layer processes an embedding sequence $X \in \mathbb{R}^{L \times d_{\text{model}}}$, where $L$ is the sequence length and $d_{\text{model}}$ is the hidden dimension. The embedding is linearly projected into $H$ distinct attention heads. For the $h$-th head, the queries (Q), keys (K), and values (V) are computed as:$$Q_h = XW_Q^h, \quad K_h = XW_K^h, \quad V_h = XW_V^h$$
The attention map $A_h$ is derived using a scaled dot-product, and is then multiplied by $V_h$ to produce the head's output $O_h$:$$O_h = A_h V_h = \text{softmax}\left(\frac{Q_h K_h^\top}{\sqrt{d_{\text{head}}}}\right) V_h$$
Finally, the outputs from all $H$ heads are concatenated and linearly projected to form the final MHA output.

\subsection{Inference Phases: Prefill and Decode}
As shown in Figure~\ref{fig:background}B, LLM inference is distinctively characterized by two sequential phases: \textit{Prefill} and \textit{Decode}.

\textbf{Prefill Phase:} This phase processes the input prompt in parallel and serves two primary purposes: (1) computing and storing the Key-Value (KV) cache for all input tokens to accelerate subsequent generation; (2) generating the first output token. This phase is typically \textit{compute-bound} due to the highly parallelizable matrix multiplications ($O(L^2)$ complexity for attention). The latency of this phase is referred to as Time-to-First-Token (TTFT).

\textbf{Decode Phase:} Following prefill, the model generates tokens auto-regressively, one by one. In each step, the model attends to all previous tokens (prompt + generated). This phase is typically \textit{memory-bound}, dominated by the cost of loading model weights and KV caches from memory.

\subsection{Prefix Caching}
Due to the quadratic computational complexity of the attention mechanism, the prefill phase remains a major bottleneck for processing long prompts. \textit{Prefix caching} \cite{lian2024Sglang,Kwon2023vllm} mitigates this by retaining the KV cache of frequently reused prefixes (e.g., system prompts or few-shot examples). When a new request shares a prefix with a cached entry, the serving engine retrieves the pre-computed KV tensors, bypassing redundant computation. However, traditional prefix caching strictly requires identical model architectures and weights, severely limiting its applicability in heterogeneous environments.

\subsection{Multi-LLM Systems}

In production environments, a ``one-size-fits-all'' model strategy is often inefficient. Multi-LLM systems \cite{wu2023autogen, hong2024metagpt, qian2024chatdev, manus2025, cursor2025, copilot2025, hu2024automated, li2024personal,zhou2026janus,mei2026coral,li2026adaserve,mei2025helix} leverage the distinct capabilities and cost-efficiency profiles of heterogeneous models to optimize performance. 

\begin{figure}[t]
    \centering
    \includegraphics[width=0.6\linewidth]{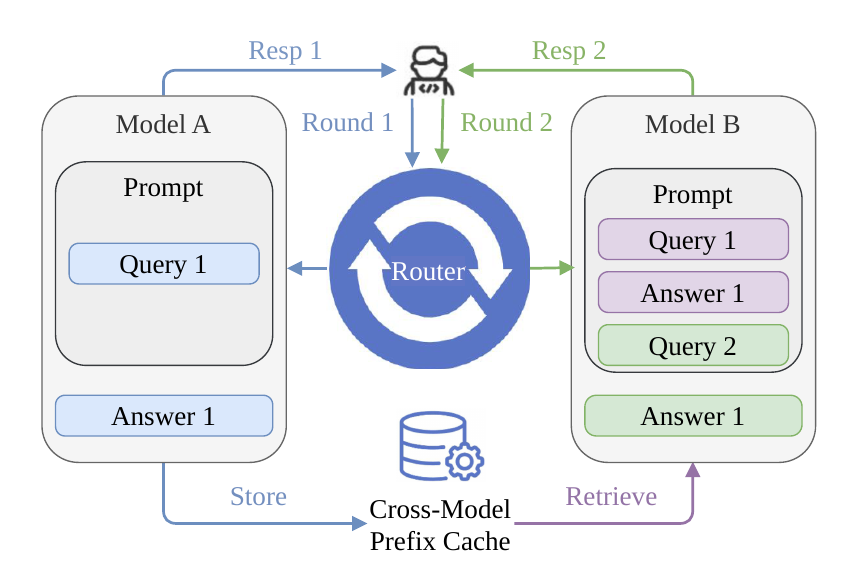}
    \caption{Illustration of cross-model prefix cache framework in multi-LLM router system.}
    \label{fig:illustration}
\end{figure}

As a typical example shown in Figure~\ref{fig:illustration}, the multi-LLM router system \cite{Ong2025routellm,chen2023frugalgpt, dekoninck2024unified} has a router that will analyze the complexity of an incoming query and direct it to the most appropriate model (e.g., sending simple queries to smaller, faster models and complex reasoning tasks to larger models). Crucially, in multi-LLM system workflows, identical prompt prefixes are often processed by different models (query 1 and answer 1 in Figure~\ref{fig:illustration}). This redundancy highlights the critical need for a cross-model prefix caching mechanism.  

\subsection{Sparse Attention}

\label{subsec:sparse_attn}
Empirical studies demonstrate that attention matrices in LLMs are highly sparse \cite{zhang2023h2o, xiao2023efficient,zhang2025pqcache,miao2024x}. Although the attention mechanism allows every token to attend to every other token, in practice, a token heavily attends to only a small subset of previous tokens.

Formally, for the $k$-th token, let the attention probability distribution be $\{P_i\}_{i=1}^k$. Given a cumulative probability threshold $p$, we define the significant attention set $S$ as the minimal set of indices that accumulate to $p$:

\begin{equation}
    S = \operatorname*{argmin}_{S \subseteq \{1, \dots, k\}} |S|, \quad \text{s.t.} \quad \sum_{i \in S} P_i \ge p
\end{equation}

Empirical results show that $|S| \ll k$, implying that the storage complexity of the attention pattern can be reduced from $O(N^2)$ to near-linear with minimal loss in precision. This structural sparsity is a key property leveraged by our proposed \sys{} to enable efficient storage and transfer of attention maps.

\reviewerthreechange[R3.M2]{Existing sparse-attention methods can be broadly divided by the inference phase they accelerate. Prefill-oriented methods, such as MInference~\cite{jiang2024minference} and FlexPrefill~\cite{lai2025flexprefill}, avoid computing the full attention matrix by identifying sparse patterns from prefill. In contrast, decode-oriented methods, such as IceCache~\cite{mao2026icecache} and ParisKV~\cite{qi2026pariskv}, first perform full prefill to construct the KV cache and exploit sparsity only during subsequent decoding. \sys{} focuses on prefill acceleration by reusing and sparsifying attention maps already computed by a sender model.}

\subsection{Cross-model Communication}
\label{sec:cross-comm}
The conventional approach for inter-model communication relies on transmitting raw token sequences, requiring the receiving model to perform a full prefill computation. While straightforward, this method incurs significant computational overhead during the prefill phase, as the context length expands rapidly over multiple rounds of interaction. To alleviate this bottleneck, recent studies leverage the intermediate KV cache, rather than raw text, as the cross-model communication medium. These techniques generally fall into two paradigms: selective partial recomputation and learned cache adaptation.

As an example of the partial recomputation paradigm, Droidspeak \cite{liu2024droidspeak} isolates critical transformer layers via profiling, selectively recomputing the KV cache only at these layers while reusing the cached tensors for the rest. Similarly, CacheBlend \cite{yao2024cacheblend} identifies tokens exhibiting significant KV state deviation between the source and target models, recomputing the KV cache exclusively for these divergent tokens. 
\reviewerthreechange[R3.W1]{
Conversely, under the learned adaptation paradigm, C2C \cite{fu2025cachetocache}  fuses the KV cache from both sender and receiver, generating a composite cache that encapsulates context from both. But it doesn't eliminate receiver prefill since it requires both models to compute the KV cache.}
\reviewerthreechange[R3.M2]{KVCOMM~\cite{ye2025kvcomm} and
Q-KVComm~\cite{kriuk2025qkvcomm} communicate or compress KV caches among
homogeneous agents whose inputs contain related content under diverging
prefixes.}
Addressing architectural heterogeneity, \cite{Lucio2026latent} introduces lightweight transformer adapters to project the sender's KV cache into an architecture-agnostic latent space, from which the receiver extracts a compatible KV representation.
\section{Motivation}
\label{sec:analysis}

\subsection{Terminology}
Table~\ref{tab:notation} summarizes the mathematical notations used in this paper.

\begin{table}[h]
  \caption{Summary of notations.}
  \label{tab:notation}
  \centering
  \begin{tabularx}{\linewidth}{{@{} l X @{}}}
    \toprule
    Notation & Description \\
    \midrule
    $S, R$ & Sender and Receiver models \\
    $L_M, H_M$ & Number of layers and heads for model $M \in \{S, R\}$ \\
    $j, i$ & Layer index ($1 \le j \le L_M$) and head index ($1 \le i \le H_M$) \\
    $r$ & Number of recomputed prefix layers in $R$ \\
    $A_i^j(M)$ & Attention map of the $i$-th head in the $j$-th layer of model $M$ \\
    $\hat{A}_i^j(M)$ & Reused (approximated) attention map for $R$ \\
    $A^{j}(M)$ & Concatenation of all heads: $[A^{j}_{1}(M), \dots, A^{j}_{H_M}(M)]$ \\
    $w^j$ & Linear projection weights for the $j$-th layer \\ 
    $m(\cdot)$ & Layer mapping function $j_R \to j_S$ \\
  \bottomrule
\end{tabularx}
\end{table}

\subsection{Challenges in Cross-Model Reuse}

Multi-LLM systems typically contain heterogeneous models with varying architectures (e.g., different layer counts, head counts and head dimensions). This heterogeneity renders naive reuse approaches, such as Droidspeak and CacheBlend, ineffective, as they are tailored strictly for homogeneous models.

To enable KV cache reuse across heterogeneous architectures, recent works have introduced learning-based approaches that employ neural networks to project KV states between models. However, these methods primarily prioritize generation quality over system efficiency. The computational overhead introduced by complex adapters often negates the latency benefits of reuse, failing to achieve the optimal accuracy-efficiency trade-off required by online serving systems.

The fundamental limitation of these approaches lies in the nature of the reused state. KV caches are \textit{tightly coupled} with model weights and hidden dimensions. Even models fine-tuned from the same base checkpoint can exhibit drastically different KV distributions for identical prompts, let alone fully heterogeneous pairs. Consequently, translating KV caches requires complex non-linear transformations to maintain accuracy, making efficient reuse intrinsically difficult.

\subsection{Attention Maps as Reuse States}
\label{sec:attn_reuse}
In contrast to KV caches, we identify the \textbf{Attention Map} as a superior candidate for cross-model state reuse. While KV states are model-specific representations, attention maps capture the \textit{semantic dependencies} between tokens (i.e., "which token attends to which"). Intuitively, if two models share the same tokenizer and context, their attention patterns should exhibit structural similarity regardless of their hidden dimensions or parameter counts.

\begin{figure}[t]
    \centering
    \includegraphics[width=0.6\linewidth]{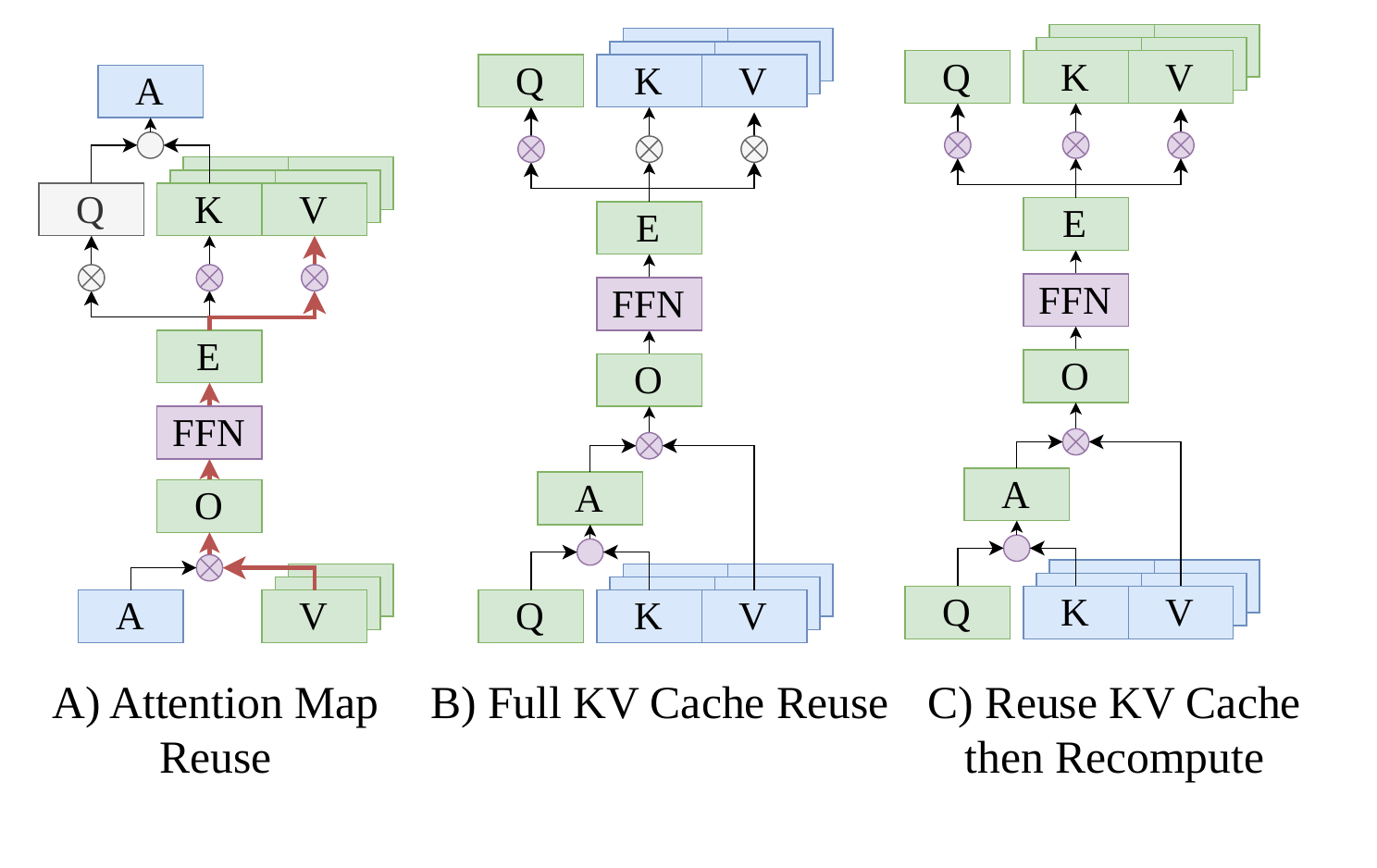}
    \caption{Illustration of attention map reuse. Each circle represents a computational module; purple circles signify modules requiring computation, while gray circles indicate inactive modules. The blue rectangles correspond to their respective caches.}
    \label{fig:attnreuse}
\end{figure}

Figure ~\ref{fig:attnreuse} shows different computation paths between attention map reuse and KV cache reuse. Compared to recomputed layers, attention map reuse avoids computing $\text{softmax}(QK^T/\sqrt{d})$ in reused layers. It must be admitted that KV cache reuse mechanism doesn't need any computation (except for last token) in full reuse layers. But the result of Droidspeak showed that a large portion of contiguous layers need recomputation to maintain accuracy, so recomputing pattern (Figure ~\ref{fig:attnreuse}) is common in KV cache reuse mechanism in practice. 
\revieweronechange[R1.M1]{Attention-map reuse also creates an opportunity to remove
key projection from the critical path of reuse layers. The receiver produces
the attention output by applying the transferred attention map to its value
vectors, so keys are not needed for that output. Key states are still required
for the subsequent decode phase, but their projection can be delayed and
overlapped with pipeline bubbles such as cross-GPU communication.}

\revieweronechange[R1.O1\\\textcolor{revisiontwo}{R2.O2}]{
Through empirical analysis, we find that certain aligned layers of a given sender–receiver model pair can exhibit an approximately linear relationship in their attention maps.
Such transferability is specific to both the model pair and the selected layer mapping, and must therefore be calibrated and validated offline rather than assumed to hold universally.}

\begin{observation}
    \label{obs:linearity}
    Given an appropriate layer mapping $m(\cdot)$, the attention map of the receiver model's layer $j$, denoted as $\{A_i^j(R)\}_{i=1}^{H_R}$, can be linearly approximated by the sender model's corresponding attention maps $\{A_i^{m(j)}(S)\}_{i=1}^{H_S}$.
\end{observation}

\reviewertwochange[R2.O3]{To identify transferable layer pairs and determine which receiver layers require recomputation, we use $L_1$ or TV (Total Variation) distance to measure the attention-map fitting error between the two models.
Specifically,
for a model pair $(R,S)$ and a target layer $j$,
we fit the following simplex-constrained regression error offline:}
\begin{equation}
\label{eq:distance}
\min_{w^j}\left\| A^{m(j)}(S) \cdot w^j - A_i^{j}(R) \right\|_1 \quad \text{s.t.} \quad w^j \ge 0, \sum w^j = 1
\end{equation}
where $w^j$ represents the linear transformation weights that aggregate sender heads to approximate a specific receiver head.
\reviewertwochange[R2.O3]{ Here we fit the weights on a small calibration set to diagnose transferability and guide layer selection. The deployed transfer weights are instead optimized using receiver’s end-to-end response loss (Section~\ref{sec:offline-training}).}

\reviewerthreechange[R3.Q1]{The simplex constraint makes each receiver head a
nonnegative, normalized convex mixture of sender heads, preserving the basic
structure of an attention distribution. This lightweight linear form can be
fused with sparse retrieval and value aggregation. Unconstrained, low-rank, or
MLP mappings could be more expressive, but would add online computation,
memory traffic, or intermediate materialization and thereby weaken the latency
benefit of reuse.}

\begin{figure*}[t]
    \centering
    \includegraphics[width=\linewidth]{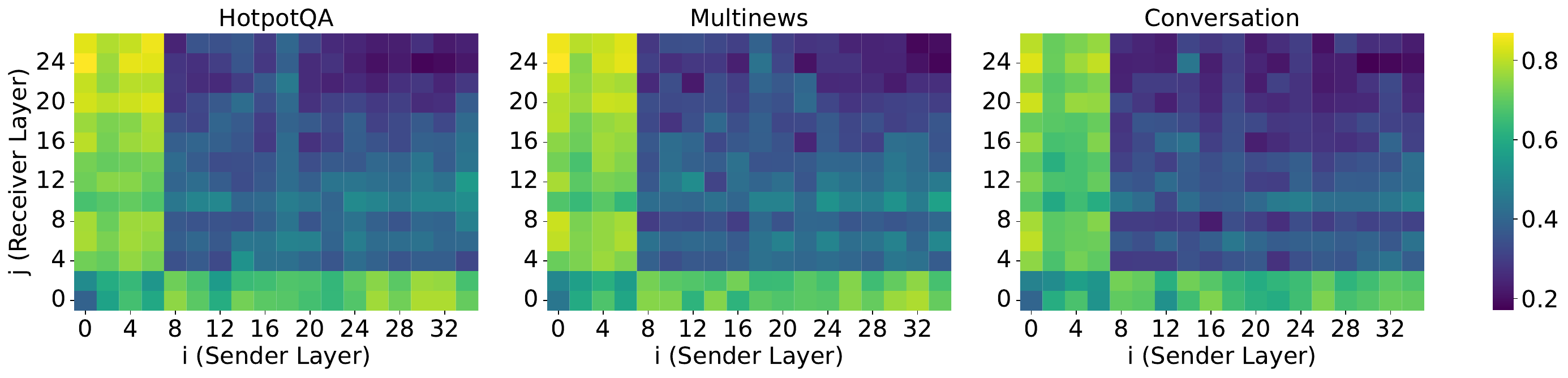}
    \caption{\revieweronechange{TV-distance heatmaps for linear fits between
    every sender--receiver layer pair in Qwen3-8B$\rightarrow$1.7B. The x-axis is the sender-layer index,
    and the y-axis is the receiver-layer index. Entry $(i,j)$ is the TV
    distance between the attention map fitted from sender layer $i$ and the
    original attention map of receiver layer $j$; lower values indicate a
    better fit.}}
    \label{fig:linear_fit}
\end{figure*}

As illustrated in Figure~\ref{fig:linear_fit}, across various datasets, \revieweronechange[R1.O1/O2/O3]{the earliest receiver layers (lower rows) exhibit high fitting errors across candidate sender layers, whereas the middle and later receiver layers can be reconstructed more accurately through linear projection.
This motivates us to recompute the first few receiver layers (e.g., $r=1 \sim 3$) and reuse attention maps thereafter. 
Our ablation on Qwen3-8B$\rightarrow$1.7B  with Conversation further confirms this trade-off: increasing $r$ from 3 to 15 provides no ROUGE-L improvement but introduces unnecessary latency (i.e., raises TTFT from 0.3205s to 0.3556s).
Using the diagnostic weights fitted above, \Sys{} profiles the layer-wise TV distance and determines $r$ offline for each model pair.
For layer mapping $m(\cdot)$, we use suffix
alignment, $m(j)=j+L_S-L_R$, where $L_S$ and $L_R$ denote the numbers of layers in the sender and receiver models, to align the empirically more stable later layers. If a model pair exhibits high fitting error, \sys{} conservatively increases r or falls back to full prefill when no reliable reuse region remains.}

\section{Data Flow in \sys{}}
\label{sec:algorithm}

Based on the observations above, we propose \sys{}, a novel LLM serving system designed to accelerate the prefill phase via attention reuse.

\begin{figure}[t]
    \centering
    \includegraphics[width=0.6\linewidth]{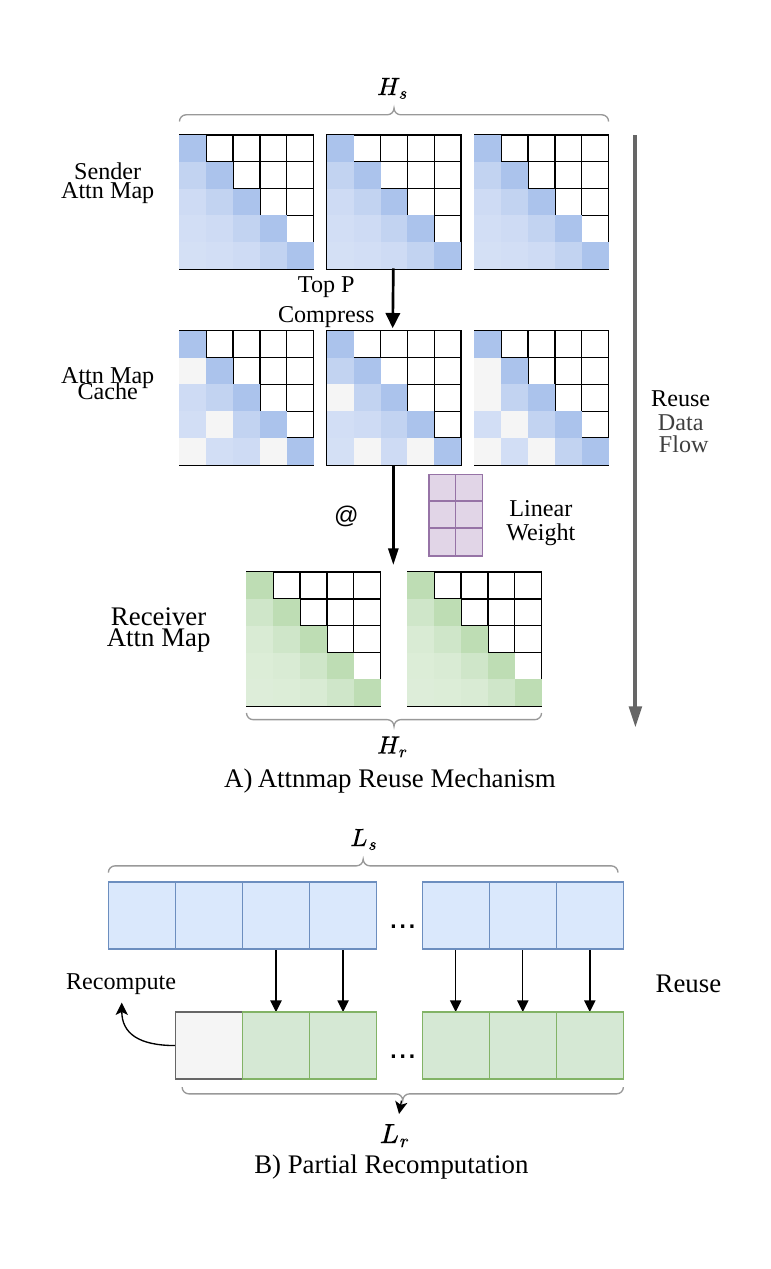}
    \caption{Overview of the data flow mechanism in \sys{}, illustrating the workflow of partial recomputation and linear attention translation.}
    \label{fig:algo_arch}
\end{figure}

As depicted in Figure~\ref{fig:algo_arch}, \sys{} contains three core components: \textit{Partial Recomputation}, \textit{Layer Mapping}, and \textit{Head Mapping}.

\textbf{Partial Recomputation}: let $r$ denote the number of recomputed layers. For the first $r$ layers, the receiver model $R$ performs standard computation to obtain accurate hidden states. For the subsequent layers ($j > r$), $R$ skips the heavy attention computation and instead reuses the cached attention maps from the sender model $S$.

\textbf{Layer Mapping}: each reuse receiver layer $j$ is mapped to a sender layer $m(j)$ and $\hat{A}^j(R)$ is sourced from $A^{m(j)}(S)$.

\textbf{Head Mapping}: Head mapping determines how each reuse receiver head $\hat{A}^j_i(R)$ is computed via $A^{m(j)}(S)$. \sys{} uses linear transformation as head mapping.

Formally, the attention map for the $i$-th head in the $j$-th layer of $R$ is derived as follows:
\begin{equation}
\label{eq:headmap}
\hat{A}_i^j(R)=
\begin{cases}
    A_i^j(R), & 1 \le j \le r \\
    A^{m(j)}(S) \cdot w^j_i, & r < j \le L_R
\end{cases}
\end{equation}
where $w^j_i$ is the learned linear transformation weight (Section~\ref{sec:offline-training}). 


\textbf{Top P Compression}: Note that reusing attention maps introduces storage challenges due to the $O(N^2)$ space complexity. It is not practical to store complete attention maps when the sequence is extremely long. We leverage sparsity in attention maps, compress them via top-$p$, and store the compressed attention maps in Block Sparse Row (BSR) format.

\textbf{Complexity Analysis}: For an input with $n$ heads, $D$ head dim and sequence length $L$, standard dense attention incurs $O(nL^2D)$ cost for attention map construction and $O(nL^2D)$ cost for applying the attention map to the value states. In contrast, \sys{} first linearly transforms the sender attention maps from $m$ heads to $n$ heads, costing $O(mnLk)$, where $k$ is the average number of retained key positions per query after top-$p$ compression. Since typically $k\ll L$, the reuse path reduces the dependence on $L$ from quadratic to near-linear. We defer the empirical characterization of $k/L$ under different top-$p$ settings to \Cref{sec:memory_footprint_analysis}.

\section{\sys{} System Design} 
\label{sec:design}

\begin{figure}[t]
    \centering
    \includegraphics[width=0.6\linewidth]{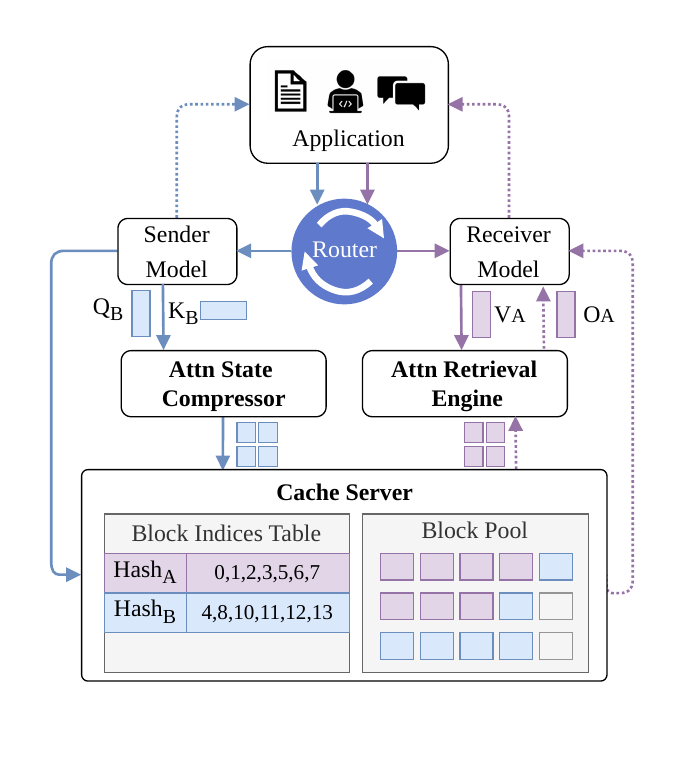}
    \caption{An overview of \sys{}}
    \label{fig:sys_overview}
\end{figure}

To efficiently reuse the attention map across different models, we must address two primary challenges. The first is mitigating memory bloat: the memory complexity of the attention map scales quadratically with the context length, making it infeasible to fully materialize it during both the training phase and runtime storage. The second challenge lies in computation: given the attention map collected from the sender, we need to efficiently apply it with transformation weights to the receiver's value vectors without incurring prohibitive inference overhead.

We design \sys{} to overcome these bottlenecks in three steps, as shown in Figure~\ref{fig:sys_overview}.
First, to reduce the memory footprint of the generated attention maps, \sys{} employs an \textit{Attention State Compressor} to compress and store the attention map in a block-wise format. 
Second, to manage the compressed attention map cache and minimize memory fragmentation, \sys{} implements a \textit{Block Manager} that organizes the cache by maintaining a prefix-hash-based indexing system for efficient retrieval. Third, to efficiently reuse the block-sparse attention map, \sys{} performs \textit{Attention Retrieval} by utilizing a custom GPU kernel which fuses the reconstruction of the receiver's attention map (using the cached sender's map and transformation weights) with the matrix multiplication of the receiver's value vectors, directly producing the attention output without intermediate dense materialization.

\subsection{\textcolor{revisionone}{Offline Calibration and Training}}
\label{sec:offline-training}

\revieweronechange[R1.O4]{For each sender--receiver model pair, \sys{} only trains the head-mapping weights (i.e., $w$ in Eq.~\eqref{eq:headmap}) while freezing both LLMs.}
\reviewertwochange[R2.O3]{The training process follows the online reusing steps. The sender prefills the shared context to produce its attention state, the receiver prefills that context through the reuse path, and the newly appended and answer tokens are then passed to the receiver with non-answer tokens masked. The mapping weights are optimized using the receiver's end-to-end response loss, rather than the TV/$L_1$ distance in \Cref{sec:attn_reuse}.}
\textcolor{revisionone}{
Note that, each reused receiver layer has
$H_S\times H_R$ scalar weights, for a total of
$(L_R-r)H_SH_R$. 
For Ministral3-8B$\rightarrow$Ministral3-3B on Conversation, the weights account for less than 0.001\% of the receiver parameters and converge within two 20-hour epochs on two A800 80GB GPUs. This one-time calibration cost can be amortized across subsequent requests, making this approach suitable for deployments with a relatively stable model pool and repeated cross-model reuse, such as multi-model routing or multi-agent serving over shared long contexts.}

\subsection{Attention State Compressor}
\label{sec:attention_compressor}
\begin{algorithm}[t]
\caption{Sparse Compression of Attention Map}
\label{alg:sparse_compress}
\begin{algorithmic}[1]
\Require Query $Q \in \mathbb{R}^{L_Q \times m \times D}$, Key $K \in \mathbb{R}^{L_{KV} \times m_{kv} \times D}$, Block size $B$, Top-$p$ threshold $p$

\Statex \textbf{Stage 1: Block-Level Top-$p$ Filtering}
\State \textbf{Parallel Execution}: Assign each threadblock to compute stats for a block pair $(q_i, k_j)$
\State $S_{i,j} \leftarrow (q_i k_j^\top) / \sqrt{D}$ \Comment{Block-level attention logits}
\State $\text{block\_row\_lse}_{i,j} \leftarrow \max_{\text{row}}(S_{i,j}) + \log \sum_{\text{row}} \exp\big(S_{i,j} - \max_{\text{row}}(S_{i,j})\big)$ \Comment{Row-wise LSE over $B \times B$ block}
\State $\text{row\_lse}_i \leftarrow \log \sum_{j} \exp(\text{block\_row\_lse}_{i,j})$ \Comment{Global row log-sum-exp}
\State $\text{block\_sumprob}_{i,j} \leftarrow \sum_{\text{col}} \exp(\text{block\_row\_lse}_{i,j} - \text{row\_lse}_i)$ \Comment{Sender block prob}
\State $\textit{mask} \leftarrow \text{Top-}p\text{\_Filter}(\text{block\_sumprob}, p)$ \Comment{Boolean mask of retained blocks}
\State $\textit{crow}, \textit{col}, \textit{nnz\_offsets} \leftarrow \text{Get\_BSR\_Metadata}(\textit{mask})$
\State Allocate $\textit{val} \in \mathbb{R}^{\text{nnz} \times B \times B}$

\Statex \textbf{Stage 2: Sparse Exact Extraction}
\State \textbf{Parallel Execution}: Assign each threadblock to process one non-zero block guided by $\textit{nnz\_offsets}$
\For{$idx = 0$ to $\text{nnz} - 1$} \Comment{Execute in parallel across non-zero blocks}
    \State $i, j, h \leftarrow \text{Decode}(\textit{nnz\_offsets}[idx])$ \Comment{Query block, Key block, Head index}
    \State $q_{block} \leftarrow \text{Load}(Q, \text{q\_idx}=i, \text{head}=h)$
    \State $k_{block} \leftarrow \text{Load}(K, \text{k\_idx}=j, \text{head}=h)$
    \State $A_{block} \leftarrow \exp(q_{block} k_{block}^\top / \sqrt{D} - \text{row\_lse}_{i,h})$ \Comment{Exact attention scores}
    \State $\textit{val}[idx] \leftarrow A_{block}$ \Comment{Store exact scores to contiguous memory}
\EndFor

\State \Return $\textit{crow}, \textit{col}, \textit{val}$
\end{algorithmic}
\end{algorithm}

\sys{} utilizes \textit{Sparse Compression} mechanism that filters the attention map before it is cached. Empirical profiling of attention distributions across extended sequence lengths reveals a heavy-tail phenomenon, as extensively studied in previous works~\cite{chen2021scatterbrain, tang2024quest,lai2025flexprefill}. This implies that retaining only the tokens necessary to satisfy a cumulative probability $p$ requires only a small fraction of the memory compared to full dense materialization, with minimal impact on generation quality. Guided by this insight, \sys{} systematically compresses the attention map by exploiting this inherent sparsity. Details about memory footprint analysis can be found at Section~\ref{sec:memory_footprint_analysis}.

Instead of applying token-level sparsity, which leads to irregular memory access patterns, \sys{} adopts a hardware-friendly block-level sparsity approach which partitions the attention map into 2D blocks (e.g., $64 \times 64$ tokens). As shown in Algorithm \ref{alg:sparse_compress}, the process avoids full dense matrix materialization by splitting the compression into two primary stages:

\paragraph{Stage 1: Block-Level Top-$p$ Filtering} 
For each query block and key block pair, we efficiently compute an aggregated block-level importance score via LogSumExp reductions (\texttt{block\_row\_lse}). Using these block-level statistics, we calculate the sum probabilities for each block (\texttt{block\_sumprob}) and apply a Top-$p$ filtering strategy. For each blocks in a row, we sort the target key blocks by their aggregated scores in descending order and retain only the subset of blocks required to reach a cumulative probability threshold $p$. This yields a binary mask indicating which blocks are to be kept.

\paragraph{Stage 2: Sparse Extraction} 
Once the block-sparsity mask is determined, we extract the BSR metadata, including compressed row indices (\texttt{crow}), column indices (\texttt{col}), and non-zero offsets (\texttt{nnz\_offsets}) which map the logical coordinates of each retained block within the $[\text{num\_blocks\_q},\allowbreak\text{num\_blocks\_k},\allowbreak\text{heads}]$ space to its physical index in the compacted 1D memory array. Guided by these offsets, a specialized kernel (\texttt{sparse\_extract}) computes the exact attention scores $\exp(Q K^\top - \text{row\_lse})$ only for the retained sparse blocks and stores them in a contiguous 1D array of dense blocks (\texttt{values}). 

This two-stage compression packs the attention map into a BSR layout ultimately, drastically reducing the memory footprint and network transmission payload and aligning with memory constraints and layout of the downstream fused kernel.
\revieweronechange[R1.O7]{
Note that, compression can be performed once after the sender produces the attention map, without blocking the sender’s critical path. And the compressed map can be reused by multiple receivers.}

\subsection{Block Manager}
\label{sec:block_manager}

\sys{} organizes the attention map cache in a block-wise structure to bound memory consumption, analogous to how vLLM manages the KV cache in paged blocks\cite{Kwon2023vllm}. However, a fundamental distinction complicates our design: whereas KV caches maintain uniform shape across layers and sequence positions, block-sparse attention maps exhibit highly dynamic shapes. Specifically, the attention context length varies across different sequence positions due to the causal masking, while the number of retained attention blocks fluctuates across different layers depending on the local sparsity patterns. This variability makes dynamic memory management more challenging.

\paragraph{Block Structure and Indexing.} To accommodate these irregular shapes, we partition the attention map into blocks along the sequence dimension. For each token span (contiguous token segment), we compute a prefix hash that incorporates the content of the span and all preceding context. This hash serves as a unique identifier for cache retrieval. We maintain, for each layer of the sender model, a list of block indices corresponding to the non-zero columns of the attention map for that token span. This sparse indexing allows \sys{} to store only the attended positions rather than the full quadratic attention matrix.

\paragraph{Cache Storage.} During the caching phase, after the sender model computes the attention map, it identifies the retained column indices for each token span (as determined by the sparsifying strategy in Section~\ref{sec:attention_compressor}). The sender transmits these indices along with their prefix hashes to the \sys{} cache server. The server maintains a hash table mapping each prefix hash to its corresponding block indices across all layers, enabling efficient lookup during retrieval.

\paragraph{Cache Retrieval and Overlap.} During online serving, the receiver model identifies the prefix hashes of the context it wishes to reuse. It transmits these hashes to the cache server, which returns the corresponding block indices for cache hits. The server then streams the hit blocks to the receiver's GPU memory. This retrieval process is fully overlapped with the receiver's partial recomputation phase described in Section~\ref{sec:algorithm}, ensuring that the attention map blocks are available before the receiver reaches the layers that require them. This overlap hides the network and I/O latency of cache transfer, preventing stalls in the generation pipeline.

\subsection{Attention Retrieval with Fused Kernel}
\label{sec:fuse}
\begin{algorithm}[t]
\caption{Fused decompress reuse attention Kernel}
\label{alg:fused_kernel}
\begin{algorithmic}[1]
\Require Sender Attention $A\in \mathbb{R}^{L_B\times L_B m}$ in BSR format (\textit{crow}, \textit{col}, \textit{val}), Receiver Values $V \in \mathbb{R}^{L_B \times n \times D}$, Weights $W \in \mathbb{R}^{m \times n}$, Block size $B$

\State \textbf{Parallel Execution}: Assign each threadblock to proceed one row of blocks ($r_{i}$) and one head of recv values ($h_{i}$)
\State Initialize accumulators: $Acc_{attn} \leftarrow 0$, $Acc_{out} \leftarrow 0$
\State $p_{start} \leftarrow \textit{crow}[r_{i}]$, $p_{end} \leftarrow \textit{crow}[r_{i} + 1]$, $last\_seq\_pos \leftarrow 0$

\For{$p = p_{start}$ to $p_{end}$} \Comment{Iterate sparse blocks}
    \State $cur\_col \leftarrow \textit{col}[p]$
    \State $cur\_seq\_pos \leftarrow cur\_col // m$
    \State $h^s \leftarrow cur\_col \bmod m$
    
    \If{$cur\_seq\_pos \neq last\_seq\_pos$} \Comment{Flush one sequence pos accumulated from different send heads}
        \State $V_{block} \leftarrow \text{Load}(V, \text{head\_idx}=h_i, \text{pos}=last\_seq\_pos)$
        \State $Acc_{out} \leftarrow Acc_{out} + \text{Matmul}(Acc_{attn}, V_{block})$
        \State $Acc_{attn} \leftarrow 0$
        \State $last\_seq\_pos \leftarrow cur\_seq\_pos$
    \EndIf

    \State Accumulate different send heads' attention of one position
    \State $A_{block} \leftarrow \textit{val}[p]$ \Comment{Load sparse attention block ($B \times B$)}
    \State $W_{scalar} \leftarrow \text{Load}(W, \text{indices based on } col)$
    \State $Acc_{attn} \leftarrow Acc_{attn} + (A_{block} \times w_{scalar})$ \Comment{Linear transform}
\EndFor

\Statex Final Flush for the last sequence block
\State $V_{block} \leftarrow \text{Load}(V, \text{head\_idx}=h_i, \text{pos}=last\_seq\_pos)$
\State $Acc_{out} \leftarrow Acc_{out} + \text{Matmul}(Acc_{attn}, V_{block})$

\State \text{Store} $Acc_{out}$ to $Output[r_{idx}]$
\end{algorithmic}
\end{algorithm}
To retrieve sender's attention map and transform into receiver's, a naive implementation would follow a two-stage pipeline: (1) materialize a dense receiver attention map $A^R \in \mathbb{R}^{L \times L \times n}$ by transforming the sender map $A^S \in \mathbb{R}^{L \times L \times m}$ with weights $W \in \mathbb{R}^{m \times n}$, and (2) multiply $A^R$ with the receiver values $V^R$ to produce the output. However, even though $A^S$ is stored in a block-sparse format, the head transformation induced by $W$ densifies attention across heads.
Consequently, explicitly materializing $A^R$ requires $\mathcal{O}(nL^2)$ memory and is infeasible for long contexts.

To overcome this memory bottleneck, \sys{} employs a custom fused GPU kernel (Algorithm~\ref{alg:fused_kernel}) that reconstructs the receiver's attention and computes the output on the fly, without allocating global memory for intermediate dense states. The attention retrieval engine organizes the sparse $A^S$ logically into a shape of $[L_B, L_B, m, B, B]$, where $L_B$ is the number of blocks along the sequence dimension and $B$ is the block size. Within the BSR layout, the column array (denoted as \textit{col} in Algorithm~\ref{alg:fused_kernel}) stores the non-zero indices of a flattened 2D logical grid of size $[L_B, m]$. Because this flattening process strictly packs the $m$ sender heads for a single sequence position contiguous in memory before advancing to the next sequence position, a 1D compressed column index (\textit{cur\_col}) inherently encodes both spatial dimensions. Therefore, during kernel execution, the original 2D coordinates can be efficiently recovered on the fly via standard integer arithmetic: integer division by $m$ yields the sequence block position (\textit{cur\_seq\_pos} $\leftarrow$ \textit{cur\_col} $/ / m$), while the modulo operation extracts the corresponding sender head index ($h^s \leftarrow$ \textit{cur\_col} $\bmod\ m$).

During computation, each threadblock is assigned to process one receiver block row $r_i$ for a specific receiver head $h_i$. The effective receiver attention is defined as $A^R_{h_i} = \sum_{k=1}^{m} W_{k, h_i} A^S_k$, and the corresponding output is $O_{h_i} = A^R_{h_i} V^R_{h_i}$. As the kernel streams over the non-zero blocks of $A^S$, the localized index layout ensures that it can incrementally accumulate the attention contributions from all $m$ sender heads for the current sequence position, weighted by the linear projection scalar $W_{k, h_i}$. Once all sender heads for a given sequence block are accumulated (detected when \textit{cur\_seq\_pos} changes), the kernel immediately executes a matrix multiplication between this aggregated attention block and the corresponding receiver value block $V^R$, accumulating the result directly into the final output register. 

This operator fusion entirely eliminates the need to instantiate $A^R$ in High Bandwidth Memory (HBM). By keeping intermediate attention scores on-chip (in registers or shared memory) throughout the computation, it drastically reduces global memory traffic. Additional implementation-level optimizations, such as cache-aware memory scheduling and autotuning configurations, are detailed in Appendix C~\cite{mapflowappendix}.

\section{Implementation}
\label{sec:implementation}

We implement the \sys{} inference server in 5K lines of code in Python, comprising three main modules: the sender, the receiver, and the cache server. We integrated our sender and receiver modules into vLLM~\cite{Kwon2023vllm}. We wrote a fused retrieval kernel using Triton~\cite{tillet2019triton}, supporting features including PagedAttention~\cite{Kwon2023vllm}, Chunked Prefill~\cite{agrawal2024chunked} and Variable Length Sequences Packing. Currently we store the compressed attention map in GPU memory for fast retrieval due to the large size of attention maps and limited bandwidth of PCIe. In order to reduce interference when multiple processes access the shared GPU resources, we launch the NVIDIA MPS server~\cite{nvidia-mps} to create a unified CUDA context for all processes and only reserve a little fraction of GPU SMs for cache server to operate inside receiver's GPU memory without interrupting the receiver's main workflow. Inter-module tensor communication leverages CUDA Inter-Process Communication (IPC)~\cite{cuda_ipc_guide}, which allows tensors to be accessed across process boundaries by passing hardware memory handles over TCP connections, eliminating the need for explicit data copies through the CPU.  

A naive IPC approach would invoke \texttt{cudaIpcOpenMemHandle} for each tensor transfer, incurring significant overhead from repeated CUDA handle operations. To mitigate this, \sys{} pre-allocates a large, fixed-size memory buffer in GPU memory and registers it for IPC during server initialization. The buffer handle is shared once across processes via TCP. Subsequently, modules write tensors to pre-negotiated offsets within this buffer and signal availability through lightweight metadata messages. This design amortizes the cost of handle creation and avoids per-transfer CUDA API overhead.

The cache server performs attention map reconstruction and compression entirely on the GPU. Since compression is applied once during the caching phase, the computational cost is independent of the number of receivers. This allows \sys{} to scale efficiently to serve multiple receiver instances without recomputing or recompressing the cached attention maps.


\reviewertwochange[R2.M1]{
Our current implementation follows prior works~\cite{li2023alpaserve,zhu2024muxserve,hong2026efficient} on multi-model multiplexing serving that assume a co-located deployment to isolate the computational benefit of attention-map reuse, but the method itself does not require models to reside on the same GPU.
In distributed settings, the compressed block-sparse attention map can be transferred to the receiver, where the fused retrieval kernel is applied locally.
}

\section{Experiments}
\label{sec:experiments}

\subsection{Setup}

\textbf{\textit{Models:}} We conduct our evaluation of \sys{} using two distinct model families: Qwen3~\cite{qwen2025qwen3}, Ministral3~\cite{liu2026ministral3}. To analyze performance across different scales, we select three model sizes and form two pairs for each family: Qwen3 (1.7B, 8B, and 14B), Ministral3 (3B, 8B, and 14B).

\noindent \textbf{\textit{Hardware Configuration:}} All experiments are conducted on virtual machines with $8 \times$ NVIDIA A800 (80GB) GPUs.

\noindent \textbf{\textit{Datasets:}} 
We evaluate \sys{} on three diverse datasets (top-$p=0.7$ by default), including HotpotQA~\cite{yang2018hotpotqa} , LMSYS~\cite{zheng2023lmsyschat1m}, ShareGPT4~\cite{chiang2023vicuna}, and Multi-News~\cite{fabbri-etal-2019-multi}.  
For each dataset, we utilize a split of approximately 10,000 examples for training and validation, and 1,000 examples for testing. Following conventions~\cite{bai2024longbench,liu2024cachegen,yao2024cacheblend}, we assess generation quality using task-specific metrics: F1 scores for Question-Answering tasks, and ROUGE-L for conversation and summarization tasks.

\noindent \textbf{\textit{Baselines:}} We compare \sys{} against state-of-the-art methods that also optimize the trade-off between generation quality and latency. Specifically, we select:
\begin{itemize}
    \item \textbf{Full Prefill:} The receiver model prefills the text of the context with vLLM~\cite{Kwon2023vllm}, representing the highest computation overhead but the best quality.
    \item \textbf{KV Cache Reuse:} We extend CacheBlend's~\cite{yao2024cacheblend} algorithm to determine the fraction of tokens to recompute for cross model KV cache reuse based on the difference between the recomputed KV cache and sender model's original KV cache. In addition, we adopt a reverse layer mapping strategy since now sender has more layers than receiver, which is different from CacheBlend's setting.
    \item \textbf{Sparse Prefill:} We also compare with FlexPrefill~\cite{lai2025flexprefill} \reviewerthreechange[R3.M2]{and MInference~\cite{jiang2024minference}} which reduce computation costs using sparse attention patterns instead of full attention.
    \item  \reviewerthreechange[R3.W1]{\textbf{Attention Map Reuse:} We compare with IAM~\cite{zhao2025iam} and implement it ourselves since its repository is empty, following its suggested 50\% layer-mapping setting.}
\end{itemize}

\noindent We did not compare with prior works with strong application constraints, like DroidSpeak using selective recomputation and necessitating embedding tensor transfers at transition layers, which is infeasible for heterogeneous model pairs with differing hidden dimensions (i.e., different shape of embedding tensors).

\begin{figure*}[t] 
    \centering
    \includegraphics[width=1\textwidth]{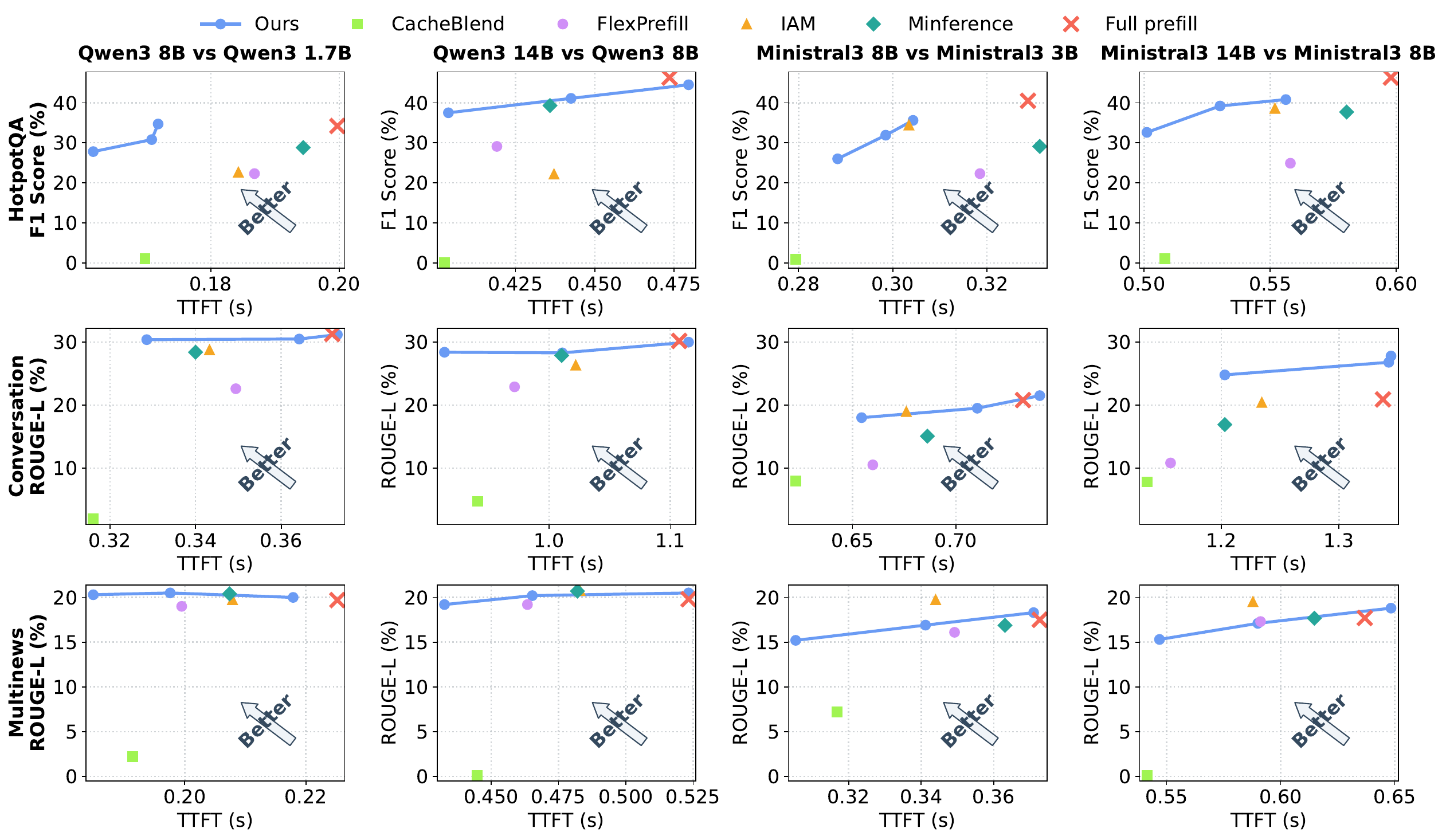} 
    \caption{\textcolor{revisionthree}{\sys{} reduces prefill latency while maintaining generation quality.}}
    \label{fig:ttft_accuracy}
\end{figure*}

\subsection{Prefill Latency–Generation Quality Trade-off}
\label{sec:exp_quality_latency}

We first demonstrate \sys{}'s favorable trade-off between TTFT reduction and generation quality in \Cref{fig:ttft_accuracy}. Across four pairs of models on three datasets, \sys{} achieves a 15\%--20\% reduction in prefill delay over the full prefill method, without compromising generation quality. Compared to FlexPrefill, \sys{} can achieve a 7\%--12\% further reduction in prefill delay with better or equal generation quality. 
\reviewerthreechange[R3.M2]{Compared to MInference, the prefill delay reduction is more significant.}
In contrast, CacheBlend suffers a significant degradation in accuracy. This outcome aligns with our analysis in \Cref{sec:analysis}, where we argue that the KV cache is tightly coupled with model-specific parameters, leading to substantial distribution differences across heterogeneous models. CacheBlend's strategy of recomputing only a fraction of tokens is insufficient to bridge this fundamental gap, resulting in a notable drop in quality.
\reviewerthreechange[R3.W1]{Compared to IAM, \sys{} provides better generation quality and better latency on most settings because of more robust and flexible attention map reusing design.}

\revieweronechange[R1.O5/O6]{The existing results also reveal model-pair and task
sensitivity. The evaluated Ministral3 pairs converge in fewer epochs and
preserve generation quality more consistently than the Qwen3 pairs.
Conversation and summarization are generally well suited to reuse because much
of a long prefix provides broad contextual grounding. QA is more sensitive:
small attention errors can change which evidence is selected. These trends
reinforce that reuse should be calibrated per model pair and workload rather
than enabled as an unchecked universal optimization.}

\begin{table}[t]
  \centering
  \caption{\textcolor{revisionthree}{F1 / TTFT (s) results on RULER-20K.}}
  \label{tab:ruler20k}
  \textcolor{revisionthree}{
  \begin{tabular}{lcccccc}
    \toprule
    Model pair & Full prefill & FlexPrefill & IAM & Minference & MapFlow\\
    \midrule
    Qwen3 8B & 0.305 / & 0.288 /  & 0.165 / & 0.133 / & 0.288 /  \\
    $\rightarrow$ 1.7B & 0.5726 & 0.5001 & 0.5285 & 0.4232 & 0.4928 \\
    \midrule
    Qwen3 14B & 0.526 / & 0.380 /  & 0.204 / & 0.259 / & 0.392 /  \\
    $\rightarrow$ 8B &  1.8604 &  1.6026 &  1.7171 &  1.3409 &  1.5767 \\
    \midrule
    Ministral3 8B & 0.454 /  & 0.318 /  & 0.344 /  & 0.176 /   & 0.348 /  \\
    $\rightarrow$ 3B & 1.2043 & 1.0530 & 1.1116 & 0.8638  & 1.0434 \\
    \midrule
    Ministral3 14B  & 0.550 /  & 0.406 /  & 0.478 /  & 0.198 /  & 0.481 /  \\
    $\rightarrow$ 8B &  2.4112 &  2.1697 &  2.2255 &  1.9719 &  2.1557 \\
    \bottomrule
  \end{tabular}
  }
\end{table}

\subsection{\textcolor{revisionthree}{Long-Context Evaluation}}
\label{sec:ruler20k}
\reviewerthreechange[\textcolor{revisionmeta}{Meta.R1}\\R3.W4/M2]{To evaluate the long-context regime, we increase the context length to 20K using RULER~\cite{hsieh2024ruler}. We tune top\_$p$ to preserve enough attention map information. Specifically, we use $p$ = 0.92, 0.87, 0.85, and 0.87 for Qwen3 8B$\rightarrow$1.7B, Qwen3 14B$\rightarrow$8B, Ministral3 8B$\rightarrow$3B, Ministral3
14B$\rightarrow$8B, respectively in \sys{}.
\sys{} achieves better F1 over FlexPrefill, IAM, and MInference, for all four pairs, while maintaining competitive prefill latency, which demonstrates the effectiveness of our approach for long context scenarios.
We also find that the latency speedups are not as significant as the results on short datasets due to larger top-$p$.
More broadly, \sys{} is complementary to receiver-side sparse prefill: the former reuses attention patterns computed by the sender, while the latter discovers sparse patterns within the receiver.
Exploring attention-map-guided hybrid strategies that combine these two mechanisms is a promising direction for improving long-context serving.}

\begin{figure*}[t]
    \centering
    \includegraphics[width=\textwidth]{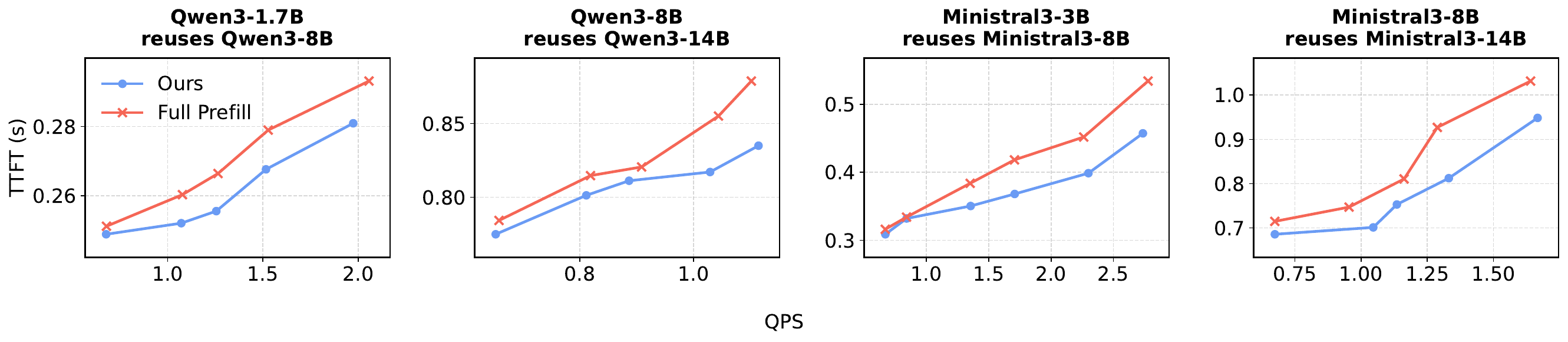}
    \caption{\sys{} achieves higher end-to-end throughput in multi-turn conversation routing.}
    \label{fig:qps_ttft}
\end{figure*}

\subsection{End-to-End Throughput Improvement on Multi-turn Conversation}

To see the impact of \sys{} on improving the throughput of an online LLM serving system with multi-model routing, we simulate the routing traces. Instead of deploying a specific learned router, we adopt the noise-injection methodology introduced in \cite{dekoninck2024unified}\cite{hu2024routerbench} to emulate prediction uncertainty. Specifically, we first evaluate the ground-truth scores of both the large and small models. We then inject zero-centered Gaussian noise into these true values to simulate the imperfect predictions of a practical router. Finally, we apply a threshold-based policy: the query is routed to the large model only if its noisy score exceeds that of the small model by a pre-defined threshold $\tau$, thereby achieving a controllable trade-off between quality and cost. For each multi-turn conversation dataset trajectory, we carefully pair each single turn with request arrival times to simulate the real-world scenario of an online system. We then run our model with optimal top-$p$ (which achieves the best performance with sufficient quality as a raw small model) and compare it with full prefill on routing.

As shown in \Cref{fig:qps_ttft}, we compare the TTFT of the receiver model under various arrival rates on the LMSYS~\cite{zheng2023lmsyschat1m} and ShareGPT4~\cite{chiang2023vicuna} dataset. \sys{} achieves lower P90 TTFT across all arrival rates, and achieves at most 25\% higher throughput (QPS) under same P90 TTFT constraint across different model pairs.

\begin{figure}[t]
    \centering
    \includegraphics[width=0.6\linewidth]{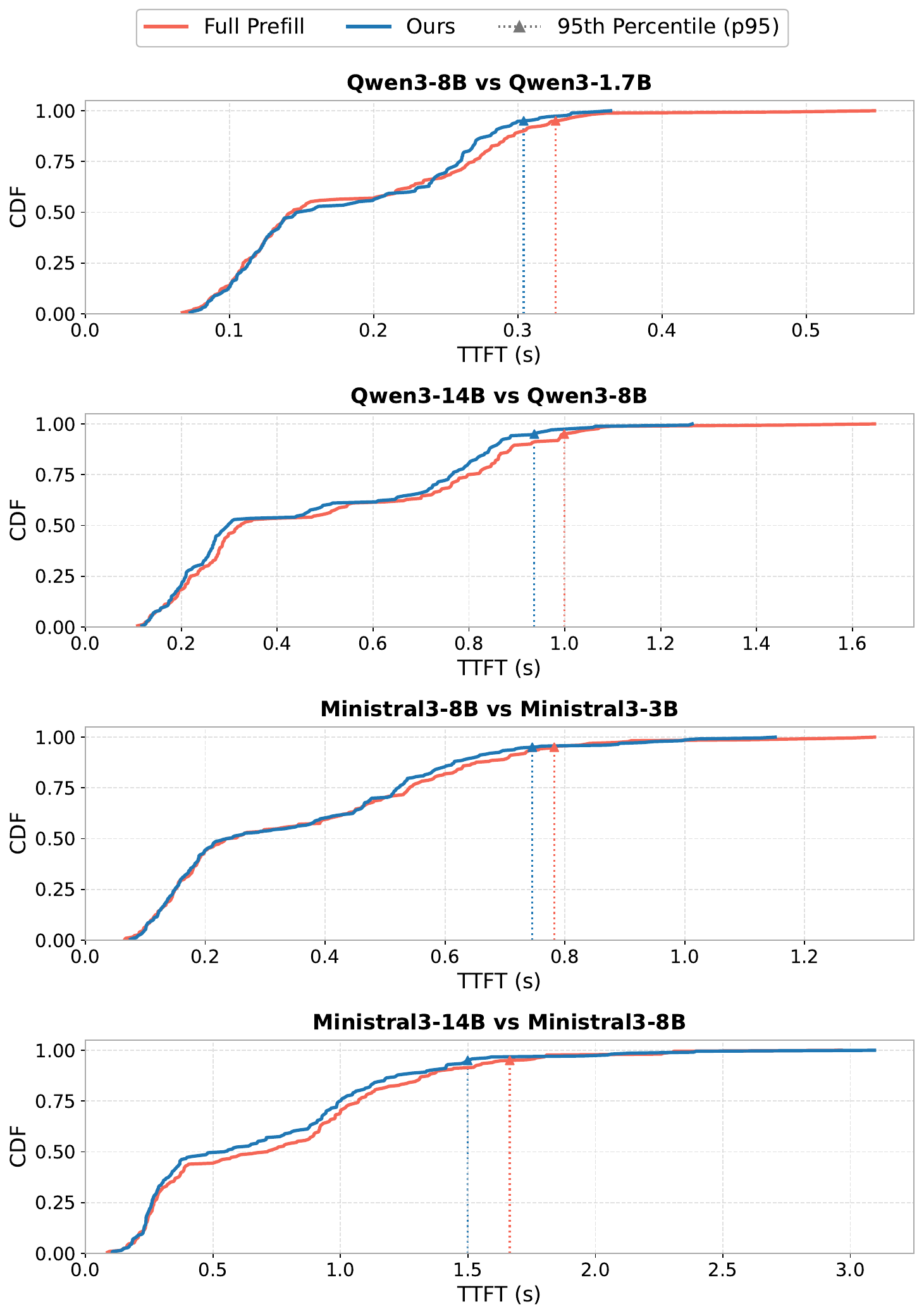}
    \caption{CDF comparison of the distribution of TTFT when switching from large to small models. The y-axis represents the cumulative probability.}
    \label{fig:large_to_small_cdf}
\end{figure}

\textbf{Understanding \sys{}'s Impact:} To further understand the impact of \sys{} on the overall system performance, we analyze the distribution of Time-To-First-Token (TTFT) for turns that are routed from the large model to the small model. \Cref{fig:large_to_small_cdf} shows the Cumulative Distribution Function (CDF) of TTFT for these turns. Across all the evaluated model pairs, the CDF curves for \sys{} consistently shift to the left compared to the full prefill baseline. More importantly, the performance gap between the two methods widens significantly at the upper tail of the distribution. In multi-turn scenarios, these tail latencies correspond to queries with extensive context histories where the quadratic complexity of standard attention becomes a severe bottleneck.

As highlighted by the 95th percentile (p95) markers, \sys{} drastically reduces tail latency. This demonstrates that reusing block-sparse attention maps effectively mitigates the compute penalty typically incurred when a small model inherits a long context from a large model. Consequently, \Sys not only accelerates average response times but ensures a much more stable and predictable user experience in online serving systems.

\subsection{Sparsification Analysis}
\label{sec:memory_footprint_analysis}

\begin{table}[htbp]
    \centering
    \caption{Memory Footprint Increments by Top\_p (GiB)}
    \label{tab:memory_footprint}
    \begin{tabular}{lccc}
        \toprule
        \textbf{Token Range} & \textbf{top\_p = 0.7} & \textbf{top\_p = 0.8} & \textbf{top\_p = 0.9} \\
        \midrule
        0$\sim$4k    & 1.23 & 1.80 & 3.11 \\
        4k$\sim$8k   & 1.84 & 3.09 & 6.35 \\
        8k$\sim$12k  & 2.82 & 4.87 & 10.07 \\
        12k$\sim$16k & 3.35 & 5.86 & 12.29 \\
        16k$\sim$20k & 2.99 & 5.53 & 12.94 \\
        20k$\sim$24k & 3.12 & 5.52 & 12.84 \\
        \bottomrule
    \end{tabular}
\end{table}

\subsubsection{Memory Footprint}
To better understand the practical efficiency of \sys{}, we analyze the memory footprint of compressed attention maps induced by block-level top-$p$ filtering across different sequence intervals. Specifically, we measure the storage cost of retained attention map blocks under varying context lengths. The following experiments are conducted
using the Qwen3-8B to Qwen3-1.7B model pair on the ShareGPT4 dataset.

As shown in \Cref{tab:memory_footprint}, the memory footprint increases initially as the sequence length grows, but it plateaus after 12k tokens. Notably, the memory usage remains stable across all top-$p$ settings in the 12k--24k token ranges. This indicates that by utilizing top-$p$ compression, \sys{} can effectively scale to long contexts without incurring additional memory overhead, making it a practical solution for real-world applications that require handling extensive conversation histories. More analysis about the effective sparsity ratio are in Appendix B.3~\cite{mapflowappendix}.

\subsubsection{\textcolor{revisiontwo}{Quality Loss}}
\label{sec:quality_loss}

\begin{table}[t]
  \centering
  \caption{\reviewertwochange{Generation quality under different
  sparsification ratios (top-$p$ thresholds) on RULER (20k context)}}
  \label{tab:sparse_loss}
  \textcolor{revisiontwo}{
  \begin{tabular}{lcccc}
    \toprule
    Model & Full attention & $p=0.9$ & $p=0.85$ & $p=0.8$ \\
    \midrule
    Qwen3 1.7B & 0.296 & 0.293 & 0.286 & 0.092 \\
    Ministral3 3B & 0.365 & 0.418 & 0.282 & 0.172 \\
    Qwen2.5 7B & 0.076 & 0.055 & 0.025 & 0.008 \\
    Llama3.1 8B & 0.531 & 0.506 & 0.306 & 0.193 \\
    Phi-4-Mini & 0.389 & 0.325 & 0.060 & 0.020 \\
    \bottomrule
  \end{tabular}
  }
\end{table}

\reviewertwochange[R2.O1]{We additionally quantify the quality loss caused by
sparsification in isolation from cross-model reuse. \Cref{tab:sparse_loss}
reports generation quality for five representative models spanning different
families under progressively more aggressive top-$p$ sparsification. Quality
degrades gracefully at top-$p=0.9$, closely tracking full attention for most
models, but drops sharply below top-$p=0.85$ for several models (e.g., Qwen3
1.7B falls from 0.286 to 0.092 between top-$p=0.85$ and $0.8$). This confirms
that the sparsification threshold cannot be fixed globally and must be
calibrated per model, consistent with \Sys's offline calibration design.}

\subsection{\textcolor{revisiontwo}{Generalizability Evaluation}}
\label{sec:exp_cross}
\begin{table}[t]
  \centering
  \caption{\reviewertwochange{Cross-dataset transfer results on 2WikiMultihopQA
  (F1 score). Weights were trained on HotpotQA and evaluated zero-shot on
  2WikiMultihopQA.}}
  \label{tab:2wiki}
  \textcolor{revisiontwo}{
  \begin{tabular}{lccc}
    \toprule
    Receiver model & Full prefill & MapFlow (top-$p$=0.9) & FlexPrefill \\
    \midrule
    Qwen3 1.7B & 0.432 & 0.335 & 0.218 \\
    Qwen3 8B & 0.464 & 0.432 & 0.226 \\
    Ministral3 3B & 0.367 & 0.403 & 0.146 \\
    Ministral3 8B & 0.513 & 0.461 & 0.223 \\
    \bottomrule
  \end{tabular}
  }
\end{table}

\reviewertwochange[R2.O2\\\textcolor{revisionthree}{R3.M1}]{
To study whether the learned transfer weights generalize beyond the calibration corpus, we perform a cross-dataset transfer generalization experiment. In particular, we train the transfer weights on HotpotQA and evaluate them zero-shot on 2WikiMultihopQA. As shown in \Cref{tab:2wiki}, generation quality remains similar without retraining, suggesting the learned weights capture model-pair-level attention correspondence rather than overfitting to a specific dataset.}

\subsection{Case Study of Different Routing Thresholds}
\begin{figure}[t]
    \centering
    \includegraphics[width=0.6\linewidth]{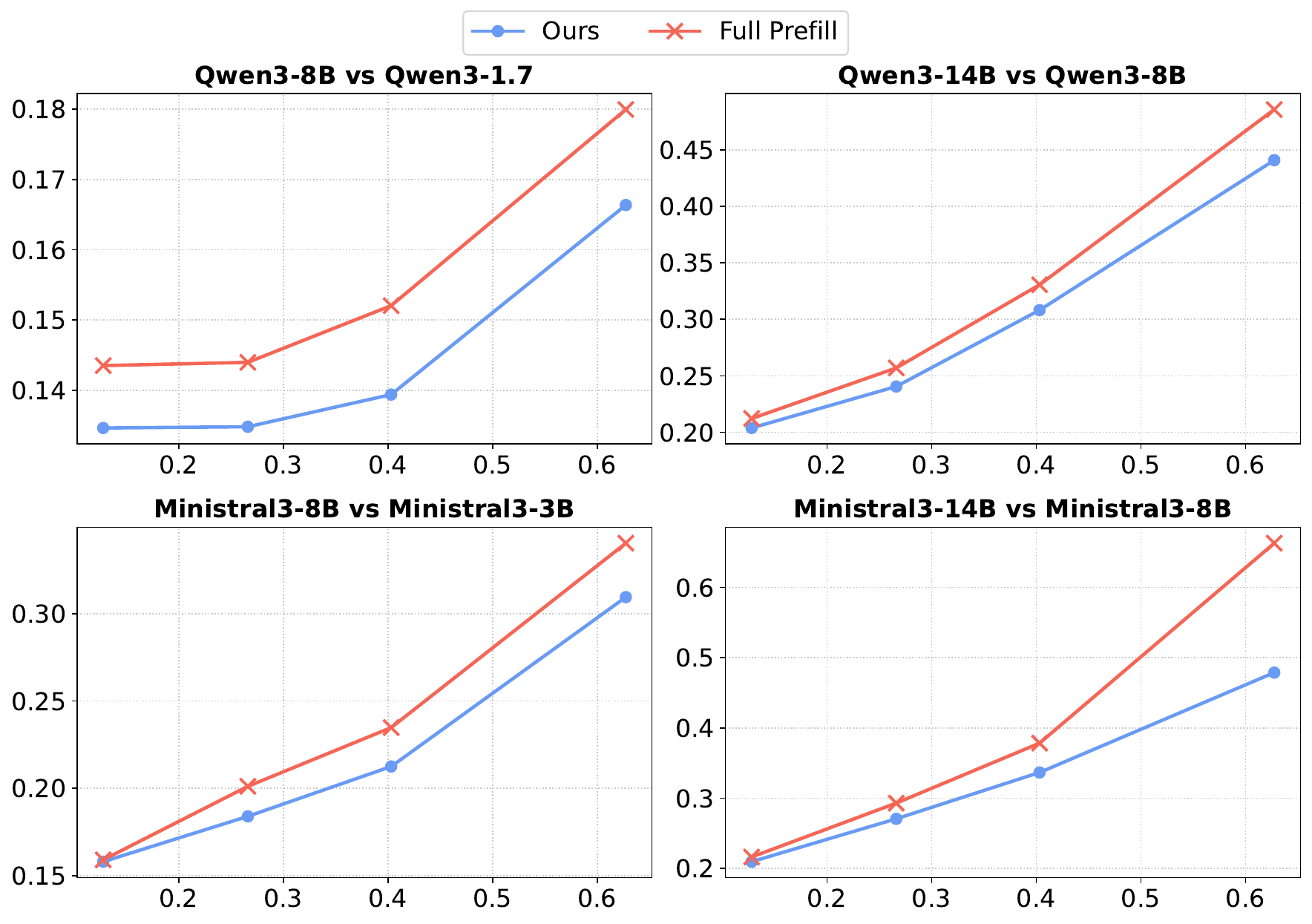}
    \caption{\sys{} achieves speedup across varying transition probabilities $p$. The x-axis represents the probability of transitioning from a large model to a small model, while the y-axis represents the P90 TTFT in seconds.}
    \label{fig:threshold}
\end{figure} 
Consider a multi-turn trajectory where a given turn $t$ ($t>1$) is routed to a small model. Let $p$ denote the probability that the preceding turn ($t-1$) was processed by a large model, meaning the probability of a consecutive small-to-small model transition is $1-p$. Furthermore, let $T_{large}$ and $T_{small}$ represent the baseline Time-To-First-Token (TTFT) for the small model when inheriting context from a large model versus a small model, respectively. Given that \sys{} achieves a cross-model prefill time ratio of $a$ ($a<1$), the expected overall TTFT speedup ratio for the system can be formulated as:

$$\text{Expected Speedup} = \frac{pT_{large} + (1-p)T_{small}}{paT_{large} + (1-p)T_{small}}$$

Rearranging this equation, we obtain:

$$\text{Expected Speedup} = \frac{1}{1-\frac{1-a}{1+(\frac{1}{p}-1)\frac{T_{small}}{T_{large}}}}$$

As $p$ increases, $T_{small}$ remains relatively constant, whereas $T_{large}$ increases significantly due to the longer contexts inherited from the large model after multiple turns, mutually contributing to a greater overall speedup. To validate this analysis, we conduct a case study by selecting different routing thresholds $\tau$ to achieve various $p$ values, measuring the P90 TTFT under a constant request arrival rate. As demonstrated in \Cref{fig:threshold}, \sys{} consistently delivers speedups across varying $p$ values. The most pronounced improvements are observed at higher $p$ values, which aligns with our theoretical expectations.

Furthermore, it is important to note that a higher $p$ indicates the system selects the large model more frequently. While this yields higher generation quality, it inherently introduces higher baseline latency. \sys{} effectively mitigates this latency penalty, enabling to maintain a favorable balance between quality and responsiveness, even under aggressive large-model routing policies.

\subsection{Single-Workflow Latency for Three Models}
\label{sec:exp_workflow}
\begin{figure}[t]
    \centering
    \includegraphics[width=0.7\linewidth]{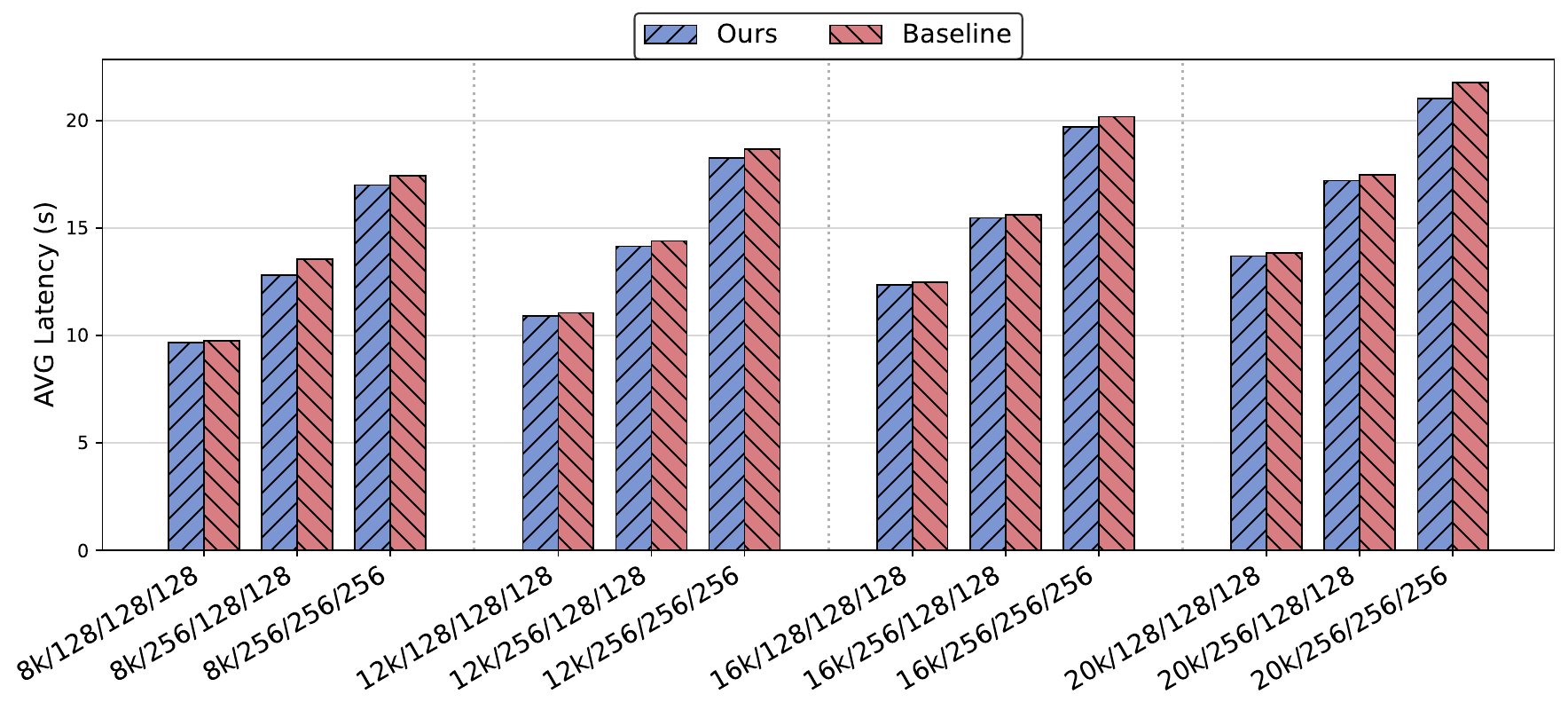}
    \caption{Speedup for single agentic workflow for Ministral model family. Horizontal axis: \textit{Prompt length} / \textit{large model decoding length} / \textit{medium model decoding length} / \textit{small model decoding length}.}
    \label{fig:three_model}
\end{figure}
We evaluate \sys{} on a representative three-model agentic
workflow~\cite{Zhanetal2025threemodel,Wangetal2025threemodel,
Sarkaretal2024threemodel}, where a request is sequentially processed by
Ministral3-14B, 8B, and 3B for planning, refinement, and summarization.
Both receiver models reuse the attention map produced by the 14B sender.
We group trajectories by context length and let each model decode a fixed
number of tokens. The 14B model uses two-way tensor parallelism, while the
8B and 3B models are co-located on one GPU to directly share the block pool.
As shown in \Cref{fig:three_model}, \sys{} consistently reduces end-to-end
latency, saving 0.48--0.73 seconds per round and achieving up to 5.4\%
speedup. This demonstrates that sender-side compression can be amortized
across multiple downstream receivers.

\reviewerthreechange[R3.W3]{This experiment demonstrates that attention-map
reuse provides a measurable end-to-end benefit in an agentic workflow, rather
than representing the maximum acceleration achievable by \sys{}. Because
\sys{} accelerates prefill but not decode, its workflow-level benefit is
naturally bounded when decoding dominates the overall latency. It is therefore
particularly suitable for TTFT-sensitive, routing-heavy, and long-context
workflows, where repeated prefill constitutes a substantial fraction of
execution time. For example, with a 20K-token context and 256 decoded tokens
across the large, medium, and small models, prefilling the two downstream
models accounts for approximately 23.4\% of the end-to-end latency. \Sys{}
accelerates these prefill stages by approximately $1.28\times$, resulting in
the observed 5.4\% end-to-end speedup.}

\section{Related Work}

\paragraph{\textbf{LLM-augmented Data Systems.}} Integrating Large Language Models (LLMs) into data management has shifted from basic natural language interfaces toward autonomous, agentic databases. Recent research explores LLM-driven database tuning and query optimization (e.g., $\lambda$-Tune~\cite{giannakouris2024lambdatune}, LLMOpt~\cite{llmopt2025}, LLM-QO~\cite{tan2025llmqo}, LlamaTune~\cite{llamatune2024}, GENO~\cite{geno2024}), where generative models replace heuristic planners to directly infer execution plans and synthesize configurations. Furthermore, frameworks increasingly leverage LLMs for complex data integration, cleaning, and entity resolution (e.g., Jellyfish~\cite{zhang2024jellyfish}, PromptEM~\cite{wang2023promptem}, Seed~\cite{seed2024}). At the orchestration level, frameworks like BridgeScope~\cite{weng2025bridgescope} and ChatDB~\cite{hu2023chatdb} securely modularize operations, enabling agents to execute ACID-compliant transactions without exposing raw data. Crucially, these LLM-augmented pipelines frequently process highly redundant prefixes—such as exhaustive database schemas, system constraints, and multi-turn instructions. Dispatching these shared contexts across heterogeneous models incurs prohibitive prefill computational costs, which directly motivate our approach for efficient cross-model state reuse mechanisms.

\paragraph{\textbf{Multi-Model Serving and Management.}} To optimize the cost-latency-accuracy tradeoff in production environments, algorithms for multi-model serving frequently employ LLM routing and cascade inference (e.g., RouteLLM~\cite{Ong2025routellm}, CascadeLLM~\cite{jiang2026cascadia}, and LLM-Blender~\cite{jiang2023llmblender}). At the system level, serving infrastructures have evolved across three paradigms. For identical model instances, high-throughput frameworks like vLLM~\cite{Kwon2023vllm} and SGLang~\cite{lian2024Sglang} utilize prefix caching to minimize redundant prefill computation. Expanding beyond a single base model, multi-adapter serving architectures—such as Punica~\cite{chen2024punica} and S-LoRA~\cite{sheng2024slora} enable the concurrent execution of thousands of fine-tuned task variants by sharing the base model's weights. Furthermore, as application workflows increasingly demand entirely distinct architectures, multi-tenant heterogeneous frameworks like AlpaServe~\cite{li2023alpaserve}, MuxServe~\cite{zhu2024muxserve}, and Llumnix~\cite{sun2024llumnix} have been developed to spatially and temporally multiplex completely different models across shared GPU clusters. All these heterogeneous serving frameworks remain unoptimized for shared prefix prompts.

\paragraph{\textbf{KV Cache Management.}} Foundational research has tackled KV Cache memory fragmentation by adapting OS-level virtualization techniques: PagedAttention enables non-contiguous memory allocation, while frameworks like SGLang~\cite{lian2024Sglang} and PromptCache~\cite{gim2024promptcache} leverage Radix Trees to facilitate prefix-aware KV cache sharing among requests directed to the same model. Concurrently, to bound cache growth during long-context generation, a proliferation of cache compression and eviction algorithms (e.g., StreamingLLM~\cite{xiao2023streamingllm}, H2O~\cite{zhang2023h2o}, SnapKV~\cite{li2024snapkv}, PQCache~\cite{zhang2025pqcache}) dynamically prune non-essential tokens by preserving only critical attention sinks or high-activation states. Despite these advancements, existing paradigms remain bound to rigid KV tensors, making them orthogonal to the attention-map-based design of \sys{}.
\section{Limitations and Future Work}
\label{sec:limit_future}

While \Sys establishes attention maps as a practical reusable state for cross-model prefix caching, our current design choices prioritize structural compatibility and online serving efficiency. We briefly discuss their implications and promising directions for future work.

\paragraph{\textbf{Applicability beyond shared tokenization.}}
\Sys is currently most effective when the sender and receiver share the same tokenizer and process the same canonicalized input context. 
\reviewertwochange[R2.O1]{This assumption holds for many realistic multi-model deployments, especially within a model family.
Taking the Llama 3 family as an example, Meta has officially released at least 15 publicly available text-model checkpoints across Llama 3, 3.1, 3.2, and 3.3\footnote{\url{https://huggingface.co/meta-llama}} that share the same underlying tokenizer vocabulary, while HuggingFace hosts hundreds of additional community-developed variants.}
There still remains many challenges to consider settings with mismatched tokenizers, prompt canonicalization pipelines, or more divergent architectures.
Extending \Sys to these broader scenarios will likely require additional alignment mechanisms, such as token-level correspondence recovery or intermediate semantic anchoring. Exploring such alignment strategies is an important direction for broadening the scope of cross-model attention reuse.


\paragraph{\textbf{Compact representations for reusable attention state.}}
Our current system adopts block-level Top-$p$ compression to make attention-map caching practical for long contexts. Although this design substantially reduces the cost of storing dense attention, reusable attention states can still occupy nontrivial memory under long-context, high-reuse workloads. This suggests an opportunity for richer compression schemes that preserve the transferability of attention structure while further reducing storage overhead. Examples include quantized sparse attention, low-rank or hybrid sparse–low-rank representations, and mechanisms that allow one sender-layer cache to serve multiple semantically aligned receiver layers. Such extensions could improve both cache capacity and reuse efficiency.

Overall, these limitations do not change the central conclusion of this work: attention maps provide a promising systems abstraction for heterogeneous prefix reuse. Instead, they highlight a broader design space for reusable cross-model execution states, which we believe will become increasingly important as multi-LLM data workflows continue to grow in complexity and diversity.

\section{Conclusion}
In this paper, we address the prefill latency bottleneck in multi-LLM systems caused by redundant prompt processing across heterogeneous models. We propose \Sys, a novel heterogeneous LLM serving system that allows different LLMs to share semantic attention maps for prefix reuse. To enable efficient cross-model sharing, we utilize a lightweight linear adapter alongside a block-level Top-$p$ sparse compression technique and a custom fused kernel to prevent memory bloat. Extensive evaluations demonstrate that \Sys achieves a 15\% to 20\% reduction in Time-to-First-Token and up to a 25\% increase in end-to-end throughput for multi-turn routing , all while maintaining generation quality. Ultimately, \Sys provides a scalable, memory-efficient solution that significantly accelerates serving times for complex multi-model workflows.

\section*{Acknowledgement}
We thank the anonymous SIGMOD reviewers for their comments and helpful feedback.
This work is supported by National Natural Science Foundation of China (U23B2048, U22B2037), Beijing Municipal Natural Science Foundation (Grant No. L2603008), and the Fundamental Research Funds for the Central Universities, Peking University. Bin Cui and Xupeng Miao are the corresponding authors.

\bibliographystyle{ACM-Reference-Format} 
\bibliography{refs} 


\begin{thebibliography}{87}


\ifx \showCODEN    \undefined \def \showCODEN     #1{\unskip}     \fi
\ifx \showISBNx    \undefined \def \showISBNx     #1{\unskip}     \fi
\ifx \showISBNxiii \undefined \def \showISBNxiii  #1{\unskip}     \fi
\ifx \showISSN     \undefined \def \showISSN      #1{\unskip}     \fi
\ifx \showLCCN     \undefined \def \showLCCN      #1{\unskip}     \fi
\ifx \shownote     \undefined \def \shownote      #1{#1}          \fi
\ifx \showarticletitle \undefined \def \showarticletitle #1{#1}   \fi
\ifx \showURL      \undefined \def \showURL       {\relax}        \fi
\providecommand\bibfield[2]{#2}
\providecommand\bibinfo[2]{#2}
\providecommand\natexlab[1]{#1}
\providecommand\showeprint[2][]{arXiv:#2}

\bibitem[Agrawal et~al\mbox{.}(2024)]%
        {agrawal2024chunked}
\bibfield{author}{\bibinfo{person}{Amey Agrawal}, \bibinfo{person}{Ashish Panwar}, \bibinfo{person}{Jayashree Mohan}, \bibinfo{person}{Nipun Kwatra}, \bibinfo{person}{Bhargav~S. Gulavani}, {and} \bibinfo{person}{Ramachandran Ramjee}.} \bibinfo{year}{2024}\natexlab{}.
\newblock \showarticletitle{SARATHI: Efficient LLM Inference by Piggybacking Decodes with Chunked Prefills}. In \bibinfo{booktitle}{\emph{18th USENIX Symposium on Operating Systems Design and Implementation (OSDI 24)}}. \bibinfo{publisher}{USENIX Association}, \bibinfo{pages}{537--552}.
\newblock


\bibitem[{Anysphere Inc.}(2025)]%
        {cursor2025}
\bibfield{author}{\bibinfo{person}{{Anysphere Inc.}}} \bibinfo{year}{2025}\natexlab{}.
\newblock \bibinfo{title}{Cursor: The AI Code Editor}.
\newblock \bibinfo{howpublished}{\url{https://www.cursor.com/}}.
\newblock
\shownote{Accessed: 2025-04-23}.
\newblock


\bibitem[Bai et~al\mbox{.}(2024)]%
        {bai2024longbench}
\bibfield{author}{\bibinfo{person}{Yushi Bai}, \bibinfo{person}{Xin Lv}, \bibinfo{person}{Jiajie Zhang}, \bibinfo{person}{Hongchang Lyu}, \bibinfo{person}{Jiankai Tang}, \bibinfo{person}{Zhidian Huang}, \bibinfo{person}{Zhengxiao Du}, \bibinfo{person}{Xiao Liu}, \bibinfo{person}{Aohan Zeng}, \bibinfo{person}{Lei Hou}, \bibinfo{person}{Yuxiao Dong}, \bibinfo{person}{Jie Tang}, {and} \bibinfo{person}{Juanzi Li}.} \bibinfo{year}{2024}\natexlab{}.
\newblock \showarticletitle{LongBench: A Bilingual, Multitask Benchmark for Long Context Understanding}. In \bibinfo{booktitle}{\emph{Proceedings of the 62nd Annual Meeting of the Association for Computational Linguistics (Volume 1: Long Papers)}}. \bibinfo{publisher}{Association for Computational Linguistics}.
\newblock


\bibitem[{Butterfly Effect AI}(2025)]%
        {manus2025}
\bibfield{author}{\bibinfo{person}{{Butterfly Effect AI}}.} \bibinfo{year}{2025}\natexlab{}.
\newblock \bibinfo{title}{Manus: AI Agent}.
\newblock \bibinfo{howpublished}{\url{https://manus.im}}.
\newblock
\shownote{Accessed: 2025-04-23}.
\newblock


\bibitem[Chen et~al\mbox{.}(2021)]%
        {chen2021scatterbrain}
\bibfield{author}{\bibinfo{person}{Beidi Chen}, \bibinfo{person}{Tri Dao}, \bibinfo{person}{Eric Winsor}, \bibinfo{person}{Zhao Song}, \bibinfo{person}{Atri Rudra}, {and} \bibinfo{person}{Christopher R{\'e}}.} \bibinfo{year}{2021}\natexlab{}.
\newblock \showarticletitle{Scatterbrain: Unifying sparse and low-rank attention}. In \bibinfo{booktitle}{\emph{Advances in Neural Information Processing Systems}}, Vol.~\bibinfo{volume}{34}. \bibinfo{pages}{17413--17426}.
\newblock


\bibitem[Chen et~al\mbox{.}(2024)]%
        {chen2024punica}
\bibfield{author}{\bibinfo{person}{Liana Chen} {et~al\mbox{.}}} \bibinfo{year}{2024}\natexlab{}.
\newblock \showarticletitle{Punica: Multi-Tenant LoRA Serving}. In \bibinfo{booktitle}{\emph{Proceedings of the 29th ACM International Conference on Architectural Support for Programming Languages and Operating Systems (ASPLOS)}}.
\newblock


\bibitem[Chen et~al\mbox{.}(2023b)]%
        {chen2023frugalgpt}
\bibfield{author}{\bibinfo{person}{Lingjiao Chen}, \bibinfo{person}{Matei Zaharia}, {and} \bibinfo{person}{James Zou}.} \bibinfo{year}{2023}\natexlab{b}.
\newblock \showarticletitle{FrugalGPT: How to Use Large Language Models While Reducing Cost and Improving Performance}. In \bibinfo{booktitle}{\emph{Advances in Neural Information Processing Systems (NeurIPS)}}.
\newblock


\bibitem[Chen et~al\mbox{.}(2023a)]%
        {seed2024}
\bibfield{author}{\bibinfo{person}{Zui Chen}, \bibinfo{person}{Lei Cao}, \bibinfo{person}{Sam Madden}, \bibinfo{person}{Tim Kraska}, \bibinfo{person}{Zeyuan Shang}, \bibinfo{person}{Ju Fan}, \bibinfo{person}{Nan Tang}, \bibinfo{person}{Zihui Gu}, \bibinfo{person}{Chunwei Liu}, {and} \bibinfo{person}{Michael Cafarella}.} \bibinfo{year}{2023}\natexlab{a}.
\newblock \showarticletitle{SEED: Domain-specific data curation with large language models}.
\newblock \bibinfo{journal}{\emph{arXiv preprint arXiv:2310.00749}} (\bibinfo{year}{2023}).
\newblock


\bibitem[Chiang et~al\mbox{.}(2023)]%
        {chiang2023vicuna}
\bibfield{author}{\bibinfo{person}{Wei-Lin Chiang}, \bibinfo{person}{Zhuohan Li}, \bibinfo{person}{Zi Lin}, \bibinfo{person}{Ying Sheng}, \bibinfo{person}{Zhanghao Wu}, \bibinfo{person}{Hao Zhang}, \bibinfo{person}{Lianmin Zheng}, \bibinfo{person}{Siyuan Zhuang}, \bibinfo{person}{Yonghao Zhuang}, \bibinfo{person}{Joseph~E. Gonzalez}, \bibinfo{person}{Ion Stoica}, {and} \bibinfo{person}{Eric~P. Xing}.} \bibinfo{year}{2023}\natexlab{}.
\newblock \bibinfo{title}{Vicuna: An Open-Source Chatbot Impressing GPT-4 with 90\%* ChatGPT Quality}.
\newblock \bibinfo{howpublished}{\url{https://lmsys.org/blog/2023-03-30-vicuna/}}.
\newblock


\bibitem[Dekoninck et~al\mbox{.}(2025)]%
        {dekoninck2024unified}
\bibfield{author}{\bibinfo{person}{Jasper Dekoninck}, \bibinfo{person}{Maximilian Baader}, {and} \bibinfo{person}{Martin~T. Vechev}.} \bibinfo{year}{2025}\natexlab{}.
\newblock \showarticletitle{A Unified Approach to Routing and Cascading for LLMs}. In \bibinfo{booktitle}{\emph{Forty-second International Conference on Machine Learning, {ICML} 2025, Vancouver, BC, Canada, July 13-19, 2025}} \emph{(\bibinfo{series}{Proceedings of Machine Learning Research})}. \bibinfo{publisher}{{PMLR} / OpenReview.net}.
\newblock
\urldef\tempurl%
\url{https://proceedings.mlr.press/v267/dekoninck25a.html}
\showURL{%
\tempurl}


\bibitem[Deng et~al\mbox{.}(2025)]%
        {deng2025alayadb}
\bibfield{author}{\bibinfo{person}{Yangshen Deng}, \bibinfo{person}{Zhengxin You}, \bibinfo{person}{Long Xiang}, \bibinfo{person}{Qilong Li}, \bibinfo{person}{Peiqi Yuan}, \bibinfo{person}{Zhaoyang Hong}, \bibinfo{person}{Yitao Zheng}, \bibinfo{person}{Wanting Li}, \bibinfo{person}{Runzhong Li}, \bibinfo{person}{Haotian Liu}, {et~al\mbox{.}}} \bibinfo{year}{2025}\natexlab{}.
\newblock \showarticletitle{AlayaDB: The Data Foundation for Efficient and Effective Long-context LLM Inference}. In \bibinfo{booktitle}{\emph{Companion of the 2025 International Conference on Management of Data}}. \bibinfo{pages}{364--377}.
\newblock


\bibitem[Dery et~al\mbox{.}(2026a)]%
        {dery2026latent}
\bibfield{author}{\bibinfo{person}{Lucio~M. Dery}, \bibinfo{person}{Zohar Yahav}, \bibinfo{person}{Henry Prior}, \bibinfo{person}{Qixuan Feng}, \bibinfo{person}{Jiajun Shen}, {and} \bibinfo{person}{Arthur Szlam}.} \bibinfo{year}{2026}\natexlab{a}.
\newblock \bibinfo{title}{Latent Space Communication via K-V Cache Alignment}.
\newblock
\showeprint[arxiv]{2601.06123}~[cs.LG]
\urldef\tempurl%
\url{https://arxiv.org/abs/2601.06123}
\showURL{%
\tempurl}


\bibitem[Dery et~al\mbox{.}(2026b)]%
        {Lucio2026latent}
\bibfield{author}{\bibinfo{person}{Lucio~M. Dery}, \bibinfo{person}{Zohar Yahav}, \bibinfo{person}{Henry Prior}, \bibinfo{person}{Qixuan Feng}, \bibinfo{person}{Jiajun Shen}, {and} \bibinfo{person}{Arthur Szlam}.} \bibinfo{year}{2026}\natexlab{b}.
\newblock \showarticletitle{Latent Space Communication via {K-V} Cache Alignment}.
\newblock \bibinfo{journal}{\emph{CoRR}}  \bibinfo{volume}{abs/2601.06123} (\bibinfo{year}{2026}).
\newblock
\href{https://doi.org/10.48550/ARXIV.2601.06123}{doi:\nolinkurl{10.48550/ARXIV.2601.06123}}


\bibitem[Dorbani et~al\mbox{.}(2025)]%
        {dorbani2025beyond}
\bibfield{author}{\bibinfo{person}{Anas Dorbani}, \bibinfo{person}{Sunny Yasser}, \bibinfo{person}{Jimmy Lin}, {and} \bibinfo{person}{Amine Mhedhbi}.} \bibinfo{year}{2025}\natexlab{}.
\newblock \showarticletitle{Beyond Quacking: Deep Integration of Language Models and RAG into DuckDB}.
\newblock \bibinfo{journal}{\emph{Proceedings of the VLDB Endowment}} \bibinfo{volume}{18}, \bibinfo{number}{12} (\bibinfo{year}{2025}), \bibinfo{pages}{5415--5418}.
\newblock


\bibitem[Fabbri et~al\mbox{.}(2019)]%
        {fabbri-etal-2019-multi}
\bibfield{author}{\bibinfo{person}{Alexander~R. Fabbri}, \bibinfo{person}{Irene Li}, \bibinfo{person}{Tianwei She}, \bibinfo{person}{Suyi Li}, {and} \bibinfo{person}{Dragomir Radev}.} \bibinfo{year}{2019}\natexlab{}.
\newblock \showarticletitle{{M}ulti-{N}ews: A Large-Scale Multi-Document Summarization Dataset and Evaluative Benchmark}. In \bibinfo{booktitle}{\emph{Proceedings of the 57th Annual Meeting of the Association for Computational Linguistics}}. \bibinfo{publisher}{Association for Computational Linguistics}, \bibinfo{pages}{1074--1084}.
\newblock
\href{https://doi.org/10.18653/v1/P19-1102}{doi:\nolinkurl{10.18653/v1/P19-1102}}


\bibitem[Fu et~al\mbox{.}(2025)]%
        {fu2025cachetocache}
\bibfield{author}{\bibinfo{person}{Tianyu Fu}, \bibinfo{person}{Zihan Min}, \bibinfo{person}{Hanling Zhang}, \bibinfo{person}{Jichao Yan}, \bibinfo{person}{Guohao Dai}, \bibinfo{person}{Wanli Ouyang}, {and} \bibinfo{person}{Yu Wang}.} \bibinfo{year}{2025}\natexlab{}.
\newblock \bibinfo{title}{Cache-to-Cache: Direct Semantic Communication Between Large Language Models}.
\newblock
\showeprint[arxiv]{2510.03215}~[cs.CL]
\urldef\tempurl%
\url{https://arxiv.org/abs/2510.03215}
\showURL{%
\tempurl}


\bibitem[Giannakouris and Trummer(2024a)]%
        {giannakouris2024demonstrating}
\bibfield{author}{\bibinfo{person}{Victor Giannakouris} {and} \bibinfo{person}{Immanuel Trummer}.} \bibinfo{year}{2024}\natexlab{a}.
\newblock \showarticletitle{Demonstrating $\lambda$-tune: Exploiting large language models for workload-adaptive database system tuning}. In \bibinfo{booktitle}{\emph{Companion of the 2024 International Conference on Management of Data}}. \bibinfo{pages}{508--511}.
\newblock


\bibitem[Giannakouris and Trummer(2024b)]%
        {giannakouris2024lambdatune}
\bibfield{author}{\bibinfo{person}{Victor Giannakouris} {and} \bibinfo{person}{Immanuel Trummer}.} \bibinfo{year}{2024}\natexlab{b}.
\newblock \showarticletitle{$\lambda$-Tune: Harnessing Large Language Models for Automated Database System Tuning}.
\newblock \bibinfo{journal}{\emph{arXiv preprint arXiv:2411.03500}} (\bibinfo{year}{2024}).
\newblock


\bibitem[Gim et~al\mbox{.}(2024)]%
        {gim2024promptcache}
\bibfield{author}{\bibinfo{person}{Insu Gim} {et~al\mbox{.}}} \bibinfo{year}{2024}\natexlab{}.
\newblock \showarticletitle{PromptCache: Modular Attention Multiplexing for LLM System}. In \bibinfo{booktitle}{\emph{Proceedings of the 29th ACM International Conference on Architectural Support for Programming Languages and Operating Systems (ASPLOS)}}.
\newblock


\bibitem[{GitHub}(2025)]%
        {copilot2025}
\bibfield{author}{\bibinfo{person}{{GitHub}}.} \bibinfo{year}{2025}\natexlab{}.
\newblock \bibinfo{title}{GitHub Copilot}.
\newblock \bibinfo{howpublished}{\url{https://github.com/features/copilot}}.
\newblock
\shownote{Accessed: 2025-04-23}.
\newblock


\bibitem[Hong et~al\mbox{.}(2024)]%
        {hong2024metagpt}
\bibfield{author}{\bibinfo{person}{Sirui Hong}, \bibinfo{person}{Mingchen Zhuge}, \bibinfo{person}{Jonathan Chen}, \bibinfo{person}{Xiawu Zheng}, \bibinfo{person}{Yuheng Cheng}, \bibinfo{person}{Jinlin Wang}, \bibinfo{person}{Ceyao Zhang}, \bibinfo{person}{Zili Wang}, \bibinfo{person}{Steven Ka~Shing Yau}, \bibinfo{person}{Zijuan Lin}, \bibinfo{person}{Liyang Zhou}, \bibinfo{person}{Chenyu Ran}, \bibinfo{person}{Lingfeng Xiao}, \bibinfo{person}{Chenglin Wu}, {and} \bibinfo{person}{J{\"{u}}rgen Schmidhuber}.} \bibinfo{year}{2024}\natexlab{}.
\newblock \showarticletitle{MetaGPT: Meta Programming for {A} Multi-Agent Collaborative Framework}. In \bibinfo{booktitle}{\emph{The Twelfth International Conference on Learning Representations, {ICLR} 2024, Vienna, Austria, May 7-11, 2024}}. \bibinfo{publisher}{OpenReview.net}.
\newblock
\urldef\tempurl%
\url{https://openreview.net/forum?id=VtmBAGCN7o}
\showURL{%
\tempurl}


\bibitem[Hong et~al\mbox{.}(2026)]%
        {hong2026efficient}
\bibfield{author}{\bibinfo{person}{Zicong Hong}, \bibinfo{person}{Yuyan Chen}, \bibinfo{person}{Haoyue Zhang}, \bibinfo{person}{Peng Li}, \bibinfo{person}{Wuhui Chen}, \bibinfo{person}{Song Guo}, {and} \bibinfo{person}{Xiaowei Shen}.} \bibinfo{year}{2026}\natexlab{}.
\newblock \showarticletitle{Efficient Multimodal Serving via Module Multiplexing}. In \bibinfo{booktitle}{\emph{Proceedings of the 21st European Conference on Computer Systems}}. \bibinfo{pages}{2292--2307}.
\newblock


\bibitem[Hsieh et~al\mbox{.}(2024)]%
        {hsieh2024ruler}
\bibfield{author}{\bibinfo{person}{Cheng-Ping Hsieh}, \bibinfo{person}{Simeng Sun}, \bibinfo{person}{Samuel Kriman}, \bibinfo{person}{Shantanu Acharya}, \bibinfo{person}{Dima Rekesh}, \bibinfo{person}{Fei Jia}, \bibinfo{person}{Yang Zhang}, {and} \bibinfo{person}{Boris Ginsburg}.} \bibinfo{year}{2024}\natexlab{}.
\newblock \showarticletitle{RULER: What's the real context size of your long-context language models?}
\newblock \bibinfo{journal}{\emph{arXiv preprint arXiv:2404.06654}} (\bibinfo{year}{2024}).
\newblock


\bibitem[Hu et~al\mbox{.}(2023)]%
        {hu2023chatdb}
\bibfield{author}{\bibinfo{person}{Chenxu Hu}, \bibinfo{person}{Jie Fu}, \bibinfo{person}{Chenzhuang Du}, \bibinfo{person}{Simian Luo}, \bibinfo{person}{Junbo Zhao}, {and} \bibinfo{person}{Hang Zhao}.} \bibinfo{year}{2023}\natexlab{}.
\newblock \showarticletitle{ChatDB: Augmenting LLMs with Databases as Their Symbolic Memory}.
\newblock \bibinfo{journal}{\emph{arXiv preprint arXiv:2306.03901}} (\bibinfo{year}{2023}).
\newblock


\bibitem[Hu et~al\mbox{.}(2024a)]%
        {hu2024routerbench}
\bibfield{author}{\bibinfo{person}{Qitian~Jason Hu}, \bibinfo{person}{Jacob Bieker}, \bibinfo{person}{Xiuyu Li}, \bibinfo{person}{Nan Jiang}, \bibinfo{person}{Benjamin Keigwin}, \bibinfo{person}{Gaurav Ranganath}, \bibinfo{person}{Kurt Keutzer}, {and} \bibinfo{person}{Shriyash~Kaustubh Upadhyay}.} \bibinfo{year}{2024}\natexlab{a}.
\newblock \bibinfo{title}{RouterBench: A Benchmark for Multi-LLM Routing System}.
\newblock
\showeprint[arxiv]{2403.12031}~[cs.ML]


\bibitem[Hu et~al\mbox{.}(2024b)]%
        {hu2024automated}
\bibfield{author}{\bibinfo{person}{Shengran Hu}, \bibinfo{person}{Cong Lu}, {and} \bibinfo{person}{Jeff Clune}.} \bibinfo{year}{2024}\natexlab{b}.
\newblock \bibinfo{title}{Automated Design of Agentic Systems}.
\newblock
\showeprint[arxiv]{2408.08435}~[cs.AI]


\bibitem[Jiang et~al\mbox{.}(2023)]%
        {jiang2023llmblender}
\bibfield{author}{\bibinfo{person}{Dongfu Jiang} {et~al\mbox{.}}} \bibinfo{year}{2023}\natexlab{}.
\newblock \showarticletitle{LLM-Blender: Ensembling Large Language Models with Pairwise Ranking and Generative Fusion}. In \bibinfo{booktitle}{\emph{Proceedings of the 61st Annual Meeting of the Association for Computational Linguistics (ACL)}}.
\newblock


\bibitem[Jiang et~al\mbox{.}(2024)]%
        {jiang2024minference}
\bibfield{author}{\bibinfo{person}{Huiqiang Jiang}, \bibinfo{person}{Yucheng Li}, \bibinfo{person}{Chengruidong Zhang}, \bibinfo{person}{Qianhui Wu}, \bibinfo{person}{Xufang Luo}, \bibinfo{person}{Surin Ahn}, \bibinfo{person}{Zhenhua Han}, \bibinfo{person}{Amir~H. Abdi}, \bibinfo{person}{Dongsheng Li}, \bibinfo{person}{Chin-Yew Lin}, \bibinfo{person}{Yuqing Yang}, {and} \bibinfo{person}{Lili Qiu}.} \bibinfo{year}{2024}\natexlab{}.
\newblock \showarticletitle{MInference 1.0: Accelerating Pre-filling for Long-Context LLMs via Dynamic Sparse Attention}. In \bibinfo{booktitle}{\emph{Advances in Neural Information Processing Systems}}.
\newblock


\bibitem[Jiang et~al\mbox{.}(2026)]%
        {jiang2026cascadia}
\bibfield{author}{\bibinfo{person}{Youhe Jiang}, \bibinfo{person}{Fangcheng Fu}, \bibinfo{person}{Wanru Zhao}, \bibinfo{person}{Stephan Rabanser}, \bibinfo{person}{Jintao Zhang}, \bibinfo{person}{Nicholas~D Lane}, {and} \bibinfo{person}{Binhang Yuan}.} \bibinfo{year}{2026}\natexlab{}.
\newblock \showarticletitle{Cascadia: An Efficient Cascade Serving System for Large Language Models}. In \bibinfo{booktitle}{\emph{The Fourteenth International Conference on Learning Representations (ICLR)}}.
\newblock


\bibitem[{Jin, Zhan and Luo, Yuze and Cui, Bin and Miao, Xupeng}(2026)]%
        {mapflowappendix}
\bibfield{author}{\bibinfo{person}{{Jin, Zhan and Luo, Yuze and Cui, Bin and Miao, Xupeng}}.} \bibinfo{year}{2026}\natexlab{}.
\newblock \bibinfo{title}{MapFlow Appendix}.
\newblock \bibinfo{howpublished}{\url{https://github.com/PKU-DAIR/MapFlow/blob/main/Appendix.pdf}}.
\newblock
\shownote{Accessed: 2028-10-10}.
\newblock


\bibitem[Kanellis et~al\mbox{.}(2022)]%
        {llamatune2024}
\bibfield{author}{\bibinfo{person}{Konstantinos Kanellis}, \bibinfo{person}{Cong Ding}, \bibinfo{person}{Brian Kroth}, \bibinfo{person}{Andreas M{\"u}ller}, \bibinfo{person}{Carlo Curino}, {and} \bibinfo{person}{Shivaram Venkataraman}.} \bibinfo{year}{2022}\natexlab{}.
\newblock \showarticletitle{LlamaTune: sample-efficient DBMS configuration tuning}.
\newblock \bibinfo{journal}{\emph{arXiv preprint arXiv:2203.05128}} (\bibinfo{year}{2022}).
\newblock


\bibitem[Kriuk and Ng(2025)]%
        {kriuk2025qkvcomm}
\bibfield{author}{\bibinfo{person}{Boris Kriuk} {and} \bibinfo{person}{Logic Ng}.} \bibinfo{year}{2025}\natexlab{}.
\newblock \showarticletitle{Q-KVComm: Efficient Multi-Agent Communication Via Adaptive KV Cache Compression}.
\newblock \bibinfo{journal}{\emph{arXiv preprint arXiv:2512.17914}} (\bibinfo{year}{2025}).
\newblock


\bibitem[Kwon et~al\mbox{.}(2023)]%
        {Kwon2023vllm}
\bibfield{author}{\bibinfo{person}{Woosuk Kwon}, \bibinfo{person}{Zhuohan Li}, \bibinfo{person}{Siyuan Zhuang}, \bibinfo{person}{Ying Sheng}, \bibinfo{person}{Lianmin Zheng}, \bibinfo{person}{Cody~Hao Yu}, \bibinfo{person}{Joseph~E Gonzalez}, \bibinfo{person}{Hao Zhang}, {and} \bibinfo{person}{Zhihao Lin}.} \bibinfo{year}{2023}\natexlab{}.
\newblock \showarticletitle{Efficient Memory Management for Large Language Model Serving with PagedAttention}. In \bibinfo{booktitle}{\emph{Proceedings of the 29th Symposium on Operating Systems Principles}}. \bibinfo{pages}{611--626}.
\newblock


\bibitem[Lai et~al\mbox{.}(2025)]%
        {lai2025flexprefill}
\bibfield{author}{\bibinfo{person}{Xunhao Lai}, \bibinfo{person}{Jianqiao Lu}, \bibinfo{person}{Yao Luo}, \bibinfo{person}{Yiyuan Ma}, {and} \bibinfo{person}{Xun Zhou}.} \bibinfo{year}{2025}\natexlab{}.
\newblock \bibinfo{title}{FlexPrefill: A Context-Aware Sparse Attention Mechanism for Efficient Long-Sequence Inference}.
\newblock
\showeprint[arxiv]{2502.20766}~[cs.LG]
\urldef\tempurl%
\url{https://arxiv.org/abs/2502.20766}
\showURL{%
\tempurl}


\bibitem[Li et~al\mbox{.}(2024b)]%
        {li2024personal}
\bibfield{author}{\bibinfo{person}{Huanchun Li}, \bibinfo{person}{Hao Wen}, \bibinfo{person}{Weijun Wang}, {et~al\mbox{.}}} \bibinfo{year}{2024}\natexlab{b}.
\newblock \bibinfo{title}{Personal LLM Agents: Insights and Survey about the Capability, Efficiency and Security}.
\newblock
\showeprint[arxiv]{2401.05459}~[cs.CL]


\bibitem[Li et~al\mbox{.}(2024a)]%
        {li2024snapkv}
\bibfield{author}{\bibinfo{person}{Yuhong Li} {et~al\mbox{.}}} \bibinfo{year}{2024}\natexlab{a}.
\newblock \showarticletitle{SnapKV: LLM Knows What You are Looking for Before Generation}.
\newblock \bibinfo{journal}{\emph{arXiv preprint arXiv:2404.14469}} (\bibinfo{year}{2024}).
\newblock


\bibitem[Li et~al\mbox{.}(2023)]%
        {li2023alpaserve}
\bibfield{author}{\bibinfo{person}{Zhuohan Li} {et~al\mbox{.}}} \bibinfo{year}{2023}\natexlab{}.
\newblock \showarticletitle{AlpaServe: Statistical Multiplexing with Foundation Models for Cloud Systems}. In \bibinfo{booktitle}{\emph{17th USENIX Symposium on Operating Systems Design and Implementation (OSDI)}}.
\newblock


\bibitem[Li et~al\mbox{.}(2026)]%
        {li2026adaserve}
\bibfield{author}{\bibinfo{person}{Zikun Li}, \bibinfo{person}{Zhuofu Chen}, \bibinfo{person}{Remi Delacourt}, \bibinfo{person}{Gabriele Oliaro}, \bibinfo{person}{Zeyu Wang}, \bibinfo{person}{Qinghan Chen}, \bibinfo{person}{Shuhuai Lin}, \bibinfo{person}{April Yang}, \bibinfo{person}{Zhihao Zhang}, \bibinfo{person}{Zhuoming Chen}, {et~al\mbox{.}}} \bibinfo{year}{2026}\natexlab{}.
\newblock \showarticletitle{Adaserve: Accelerating multi-slo llm serving with slo-customized speculative decoding}. In \bibinfo{booktitle}{\emph{Proceedings of the 21st European Conference on Computer Systems}}. \bibinfo{pages}{36--54}.
\newblock


\bibitem[Liu et~al\mbox{.}(2026)]%
        {liu2026ministral3}
\bibfield{author}{\bibinfo{person}{Alexander~H. Liu}, \bibinfo{person}{Kartik Khandelwal}, \bibinfo{person}{Sandeep Subramanian}, \bibinfo{person}{Victor Jouault}, \bibinfo{person}{Abhinav Rastogi}, {et~al\mbox{.}}} \bibinfo{year}{2026}\natexlab{}.
\newblock \bibinfo{title}{Ministral 3}.
\newblock
\showeprint[arxiv]{2601.08584}~[cs.CL]
\urldef\tempurl%
\url{https://arxiv.org/abs/2601.08584}
\showURL{%
\tempurl}


\bibitem[Liu et~al\mbox{.}(2024a)]%
        {liu2024droidspeak}
\bibfield{author}{\bibinfo{person}{Yuhan Liu}, \bibinfo{person}{Esha Choukse}, \bibinfo{person}{Shan Lu}, \bibinfo{person}{Junchen Jiang}, {and} \bibinfo{person}{Madan Musuvathi}.} \bibinfo{year}{2024}\natexlab{a}.
\newblock \showarticletitle{DroidSpeak: Enhancing Cross-LLM Communication}.
\newblock \bibinfo{journal}{\emph{CoRR}}  \bibinfo{volume}{abs/2411.02820} (\bibinfo{year}{2024}).
\newblock
\href{https://doi.org/10.48550/ARXIV.2411.02820}{doi:\nolinkurl{10.48550/ARXIV.2411.02820}}


\bibitem[Liu et~al\mbox{.}(2024b)]%
        {liu2024cachegen}
\bibfield{author}{\bibinfo{person}{Yuhan Liu}, \bibinfo{person}{Hanchen Li}, \bibinfo{person}{Yihua Cheng}, \bibinfo{person}{Siddhant Ray}, \bibinfo{person}{Yuyang Huang}, \bibinfo{person}{Qizheng Zhang}, \bibinfo{person}{Kuntai Du}, \bibinfo{person}{Jiayi Yao}, \bibinfo{person}{Shan Lu}, \bibinfo{person}{Ganesh Ananthanarayanan}, \bibinfo{person}{Michael Maire}, \bibinfo{person}{Henry Hoffmann}, \bibinfo{person}{Ari Holtzman}, {and} \bibinfo{person}{Junchen Jiang}.} \bibinfo{year}{2024}\natexlab{b}.
\newblock \showarticletitle{CacheGen: KV Cache Compression and Streaming for Fast Large Language Model Serving}. In \bibinfo{booktitle}{\emph{Proceedings of the ACM SIGCOMM 2024 Conference}}. \bibinfo{publisher}{Association for Computing Machinery}.
\newblock


\bibitem[Mao et~al\mbox{.}(2026)]%
        {mao2026icecache}
\bibfield{author}{\bibinfo{person}{Yuzhen Mao}, \bibinfo{person}{Qitong Wang}, \bibinfo{person}{Martin Ester}, {and} \bibinfo{person}{Ke Li}.} \bibinfo{year}{2026}\natexlab{}.
\newblock \showarticletitle{IceCache: Memory-efficient KV-cache Management for Long-Sequence LLMs}. In \bibinfo{booktitle}{\emph{The Fourteenth International Conference on Learning Representations}}.
\newblock


\bibitem[Mei et~al\mbox{.}(2026)]%
        {mei2026coral}
\bibfield{author}{\bibinfo{person}{Yixuan Mei}, \bibinfo{person}{Zikun Li}, \bibinfo{person}{Zixuan Chen}, \bibinfo{person}{Shiqi Pan}, \bibinfo{person}{Mengdi Wu}, \bibinfo{person}{Xupeng Miao}, \bibinfo{person}{Zhihao Jia}, {and} \bibinfo{person}{KV Rashmi}.} \bibinfo{year}{2026}\natexlab{}.
\newblock \showarticletitle{Coral: Cost-Efficient Multi-LLM Serving over Heterogeneous Cloud GPUs}.
\newblock \bibinfo{journal}{\emph{arXiv preprint arXiv:2605.04357}} (\bibinfo{year}{2026}).
\newblock


\bibitem[Mei et~al\mbox{.}(2025)]%
        {mei2025helix}
\bibfield{author}{\bibinfo{person}{Yixuan Mei}, \bibinfo{person}{Yonghao Zhuang}, \bibinfo{person}{Xupeng Miao}, \bibinfo{person}{Juncheng Yang}, \bibinfo{person}{Zhihao Jia}, {and} \bibinfo{person}{Rashmi Vinayak}.} \bibinfo{year}{2025}\natexlab{}.
\newblock \showarticletitle{Helix: Serving large language models over heterogeneous gpus and network via max-flow}. In \bibinfo{booktitle}{\emph{Proceedings of the 30th ACM International Conference on Architectural Support for Programming Languages and Operating Systems, Volume 1}}. \bibinfo{pages}{586--602}.
\newblock


\bibitem[Miao et~al\mbox{.}(2026)]%
        {DBLP:journals/csur/MiaoOZCJCJ26}
\bibfield{author}{\bibinfo{person}{Xupeng Miao}, \bibinfo{person}{Gabriele Oliaro}, \bibinfo{person}{Zhihao Zhang}, \bibinfo{person}{Xinhao Cheng}, \bibinfo{person}{Hongyi Jin}, \bibinfo{person}{Tianqi Chen}, {and} \bibinfo{person}{Zhihao Jia}.} \bibinfo{year}{2026}\natexlab{}.
\newblock \showarticletitle{Towards Efficient Generative Large Language Model Serving: {A} Survey from Algorithms to Systems}.
\newblock \bibinfo{journal}{\emph{{ACM} Comput. Surv.}} \bibinfo{volume}{58}, \bibinfo{number}{1} (\bibinfo{year}{2026}), \bibinfo{pages}{15:1--15:37}.
\newblock
\href{https://doi.org/10.1145/3754448}{doi:\nolinkurl{10.1145/3754448}}


\bibitem[Miao et~al\mbox{.}(2024a)]%
        {miao2024specinfer}
\bibfield{author}{\bibinfo{person}{Xupeng Miao}, \bibinfo{person}{Gabriele Oliaro}, \bibinfo{person}{Zhihao Zhang}, \bibinfo{person}{Xinhao Cheng}, \bibinfo{person}{Zeyu Wang}, \bibinfo{person}{Zhengxin Zhang}, \bibinfo{person}{Rae Ying~Yee Wong}, \bibinfo{person}{Alan Zhu}, \bibinfo{person}{Lijie Yang}, \bibinfo{person}{Xiaoxiang Shi}, {et~al\mbox{.}}} \bibinfo{year}{2024}\natexlab{a}.
\newblock \showarticletitle{Specinfer: Accelerating large language model serving with tree-based speculative inference and verification}. In \bibinfo{booktitle}{\emph{Proceedings of the 29th ACM International Conference on Architectural Support for Programming Languages and Operating Systems, Volume 3}}. \bibinfo{pages}{932--949}.
\newblock


\bibitem[Miao et~al\mbox{.}(2024b)]%
        {miao2024spotserve}
\bibfield{author}{\bibinfo{person}{Xupeng Miao}, \bibinfo{person}{Chunan Shi}, \bibinfo{person}{Jiangfei Duan}, \bibinfo{person}{Xiaoli Xi}, \bibinfo{person}{Dahua Lin}, \bibinfo{person}{Bin Cui}, {and} \bibinfo{person}{Zhihao Jia}.} \bibinfo{year}{2024}\natexlab{b}.
\newblock \showarticletitle{Spotserve: Serving generative large language models on preemptible instances}. In \bibinfo{booktitle}{\emph{Proceedings of the 29th ACM International Conference on Architectural Support for Programming Languages and Operating Systems, Volume 2}}. \bibinfo{pages}{1112--1127}.
\newblock


\bibitem[Miao et~al\mbox{.}(2024c)]%
        {miao2024x}
\bibfield{author}{\bibinfo{person}{Xupeng Miao}, \bibinfo{person}{Shenhan Zhu}, \bibinfo{person}{Fangcheng Fu}, \bibinfo{person}{Ziyu Guo}, \bibinfo{person}{Zhi Yang}, \bibinfo{person}{Yaofeng Tu}, \bibinfo{person}{Zhihao Jia}, {and} \bibinfo{person}{Bin Cui}.} \bibinfo{year}{2024}\natexlab{c}.
\newblock \showarticletitle{X-former Elucidator: Reviving Efficient Attention for Long Context Language Modeling.}. In \bibinfo{booktitle}{\emph{IJCAI}}. \bibinfo{pages}{8179--8187}.
\newblock


\bibitem[{NVIDIA Corporation}(2024)]%
        {cuda_ipc_guide}
\bibfield{author}{\bibinfo{person}{{NVIDIA Corporation}}.} \bibinfo{year}{2024}\natexlab{}.
\newblock \bibinfo{booktitle}{\emph{CUDA C++ Programming Guide}}.
\newblock
\shownote{Accessed: 2024-05-14}.
\newblock
\urldef\tempurl%
\url{https://docs.nvidia.com/cuda/cuda-c-programming-guide/index.html#interprocess-communication}
\showURL{%
\tempurl}


\bibitem[{NVIDIA Corporation}(2026)]%
        {nvidia-mps}
\bibfield{author}{\bibinfo{person}{{NVIDIA Corporation}}.} \bibinfo{year}{2026}\natexlab{}.
\newblock \bibinfo{booktitle}{\emph{Multi-Process Service ({MPS})}}.
\newblock NVIDIA.
\newblock
\shownote{Accessed: 2026-04-17}.
\newblock
\urldef\tempurl%
\url{https://docs.nvidia.com/deploy/mps/index.html}
\showURL{%
\tempurl}


\bibitem[Ong et~al\mbox{.}(2025)]%
        {Ong2025routellm}
\bibfield{author}{\bibinfo{person}{Isaac Ong}, \bibinfo{person}{Amjad Almahairi}, \bibinfo{person}{Vincent Wu}, \bibinfo{person}{Wei{-}Lin Chiang}, \bibinfo{person}{Tianhao Wu}, \bibinfo{person}{Joseph~E. Gonzalez}, \bibinfo{person}{M.~Waleed Kadous}, {and} \bibinfo{person}{Ion Stoica}.} \bibinfo{year}{2025}\natexlab{}.
\newblock \showarticletitle{RouteLLM: Learning to Route LLMs from Preference Data}. In \bibinfo{booktitle}{\emph{The Thirteenth International Conference on Learning Representations, {ICLR} 2025, Singapore, April 24-28, 2025}}. \bibinfo{publisher}{OpenReview.net}.
\newblock
\urldef\tempurl%
\url{https://openreview.net/forum?id=8sSqNntaMr}
\showURL{%
\tempurl}


\bibitem[Qi et~al\mbox{.}(2026)]%
        {qi2026pariskv}
\bibfield{author}{\bibinfo{person}{Yanlin Qi}, \bibinfo{person}{Xinhang Chen}, \bibinfo{person}{Huiqiang Jiang}, \bibinfo{person}{Qitong Wang}, \bibinfo{person}{Botao Peng}, {and} \bibinfo{person}{Themis Palpanas}.} \bibinfo{year}{2026}\natexlab{}.
\newblock \showarticletitle{ParisKV: Fast and Drift-Robust KV-Cache Retrieval for Long-Context LLMs}. In \bibinfo{booktitle}{\emph{International Conference on Machine Learning}}.
\newblock


\bibitem[Qian et~al\mbox{.}(2024)]%
        {qian2024chatdev}
\bibfield{author}{\bibinfo{person}{Chen Qian}, \bibinfo{person}{Wei Liu}, \bibinfo{person}{Hongzhang Liu}, \bibinfo{person}{Nuo Chen}, \bibinfo{person}{Yufan Dang}, \bibinfo{person}{Jiahao Li}, \bibinfo{person}{Cheng Yang}, \bibinfo{person}{Weize Chen}, \bibinfo{person}{Yusheng Su}, \bibinfo{person}{Xin Cong}, \bibinfo{person}{Juyuan Xu}, \bibinfo{person}{Dahai Li}, \bibinfo{person}{Zhiyuan Liu}, {and} \bibinfo{person}{Maosong Sun}.} \bibinfo{year}{2024}\natexlab{}.
\newblock \showarticletitle{ChatDev: Communicative Agents for Software Development}. In \bibinfo{booktitle}{\emph{Proceedings of the 62nd Annual Meeting of the Association for Computational Linguistics (Volume 1: Long Papers), {ACL} 2024, Bangkok, Thailand, August 11-16, 2024}}, \bibfield{editor}{\bibinfo{person}{Lun{-}Wei Ku}, \bibinfo{person}{Andre Martins}, {and} \bibinfo{person}{Vivek Srikumar}} (Eds.). \bibinfo{publisher}{Association for Computational Linguistics}, \bibinfo{pages}{15174--15186}.
\newblock
\href{https://doi.org/10.18653/V1/2024.ACL-LONG.810}{doi:\nolinkurl{10.18653/V1/2024.ACL-LONG.810}}


\bibitem[{Qwen Team}(2025)]%
        {qwen2025qwen3}
\bibfield{author}{\bibinfo{person}{{Qwen Team}}.} \bibinfo{year}{2025}\natexlab{}.
\newblock \bibinfo{title}{Qwen3 Technical Report}.
\newblock
\showeprint[arxiv]{2505.09388}~[cs.CL]
\urldef\tempurl%
\url{https://arxiv.org/abs/2505.09388}
\showURL{%
\tempurl}


\bibitem[Sarkar et~al\mbox{.}(2024)]%
        {Sarkaretal2024threemodel}
\bibfield{author}{\bibinfo{person}{Souvika Sarkar}, \bibinfo{person}{Mohammad~Fakhruddin Babar}, \bibinfo{person}{Monowar Hasan}, {and} \bibinfo{person}{Shubhra~Kanti Karmaker}.} \bibinfo{year}{2024}\natexlab{}.
\newblock \showarticletitle{LLMs as On-demand Customizable Service}.
\newblock \bibinfo{journal}{\emph{arXiv}} (\bibinfo{year}{2024}).
\newblock
\href{https://doi.org/10.48550/arxiv.2401.16577}{doi:\nolinkurl{10.48550/arxiv.2401.16577}}


\bibitem[Sheng et~al\mbox{.}(2024)]%
        {sheng2024slora}
\bibfield{author}{\bibinfo{person}{Ying Sheng} {et~al\mbox{.}}} \bibinfo{year}{2024}\natexlab{}.
\newblock \showarticletitle{S-LoRA: Serving Thousands of Concurrent LoRA Adapters}. In \bibinfo{booktitle}{\emph{7th Conference on Machine Learning and Systems (MLSys)}}.
\newblock


\bibitem[Shi et~al\mbox{.}(2026)]%
        {shi2026multi}
\bibfield{author}{\bibinfo{person}{Chunan Shi}, \bibinfo{person}{Yilei Chen}, \bibinfo{person}{Yilin Chen}, \bibinfo{person}{Xupeng Miao}, {and} \bibinfo{person}{Bin Cui}.} \bibinfo{year}{2026}\natexlab{}.
\newblock \showarticletitle{Multi-Segment Attention: Enabling Efficient KV-Cache Management for Faster Large Language Model Serving}.
\newblock \bibinfo{journal}{\emph{arXiv preprint arXiv:2606.02964}} (\bibinfo{year}{2026}).
\newblock


\bibitem[Singh et~al\mbox{.}(2025)]%
        {geno2024}
\bibfield{author}{\bibinfo{person}{Aditi Singh}, \bibinfo{person}{Akash Shetty}, \bibinfo{person}{Abul Ehtesham}, \bibinfo{person}{Saket Kumar}, {and} \bibinfo{person}{Tala~Talaei Khoei}.} \bibinfo{year}{2025}\natexlab{}.
\newblock \showarticletitle{A survey of large language model-based generative ai for text-to-sql: Benchmarks, applications, use cases, and challenges}. In \bibinfo{booktitle}{\emph{2025 IEEE 15th Annual Computing and Communication Workshop and Conference (CCWC)}}. IEEE, \bibinfo{pages}{00015--00021}.
\newblock


\bibitem[Sun et~al\mbox{.}(2024)]%
        {sun2024llumnix}
\bibfield{author}{\bibinfo{person}{Biao Sun} {et~al\mbox{.}}} \bibinfo{year}{2024}\natexlab{}.
\newblock \showarticletitle{Llumnix: Dynamic Scheduling for Large Language Model Serving}. In \bibinfo{booktitle}{\emph{18th USENIX Symposium on Operating Systems Design and Implementation (OSDI)}}.
\newblock


\bibitem[Tan et~al\mbox{.}(2025)]%
        {tan2025llmqo}
\bibfield{author}{\bibinfo{person}{Jie Tan} {et~al\mbox{.}}} \bibinfo{year}{2025}\natexlab{}.
\newblock \showarticletitle{Can Large Language Models Be Query Optimizer for Relational Databases?}
\newblock \bibinfo{journal}{\emph{Proceedings of the ACM on Management of Data (SIGMOD)}} \bibinfo{volume}{3}, \bibinfo{number}{6} (\bibinfo{year}{2025}), \bibinfo{pages}{306:1--306:27}.
\newblock


\bibitem[Tang et~al\mbox{.}(2024)]%
        {tang2024quest}
\bibfield{author}{\bibinfo{person}{Jiaming Tang}, \bibinfo{person}{Yilong Zhao}, \bibinfo{person}{Kan Zhu}, \bibinfo{person}{Guangxuan Xiao}, \bibinfo{person}{Baris Kasikci}, {and} \bibinfo{person}{Song Han}.} \bibinfo{year}{2024}\natexlab{}.
\newblock \bibinfo{title}{Quest: Query-Aware Sparsity for Efficient Long-Context LLM Inference}.
\newblock
\showeprint[arxiv]{2406.10774}~[cs.CL]


\bibitem[Tillet et~al\mbox{.}(2019)]%
        {tillet2019triton}
\bibfield{author}{\bibinfo{person}{Philippe Tillet}, \bibinfo{person}{H.~T. Kung}, {and} \bibinfo{person}{David Cox}.} \bibinfo{year}{2019}\natexlab{}.
\newblock \showarticletitle{Triton: an intermediate language and compiler for tiled neural network computations}. In \bibinfo{booktitle}{\emph{Proceedings of the 3rd ACM SIGPLAN International Workshop on Machine Learning and Programming Languages}}. \bibinfo{publisher}{ACM}, \bibinfo{pages}{10--19}.
\newblock


\bibitem[Vaswani et~al\mbox{.}(2017)]%
        {vaswani2017attention}
\bibfield{author}{\bibinfo{person}{Ashish Vaswani}, \bibinfo{person}{Noam Shazeer}, \bibinfo{person}{Niki Parmar}, \bibinfo{person}{Jakob Uszkoreit}, \bibinfo{person}{Llion Jones}, \bibinfo{person}{Aidan~N. Gomez}, \bibinfo{person}{Lukasz Kaiser}, {and} \bibinfo{person}{Illia Polosukhin}.} \bibinfo{year}{2017}\natexlab{}.
\newblock \showarticletitle{Attention is All you Need}. In \bibinfo{booktitle}{\emph{Advances in Neural Information Processing Systems 30: Annual Conference on Neural Information Processing Systems 2017, December 4-9, 2017, Long Beach, CA, {USA}}}. \bibinfo{pages}{5998--6008}.
\newblock
\urldef\tempurl%
\url{https://proceedings.neurips.cc/paper/2017/hash/3f5ee243547dee91fbd053c1c4a845aa-Abstract.html}
\showURL{%
\tempurl}


\bibitem[Wadlom et~al\mbox{.}(2026)]%
        {wadlom2026efficient}
\bibfield{author}{\bibinfo{person}{Noppanat Wadlom}, \bibinfo{person}{Junyi Shen}, {and} \bibinfo{person}{Yao Lu}.} \bibinfo{year}{2026}\natexlab{}.
\newblock \showarticletitle{Efficient LLM serving for agentic workflows: A data systems perspective}.
\newblock \bibinfo{journal}{\emph{arXiv preprint arXiv:2603.16104}} (\bibinfo{year}{2026}).
\newblock


\bibitem[Wang et~al\mbox{.}(2025b)]%
        {Wangetal2025threemodel}
\bibfield{author}{\bibinfo{person}{Fali Wang}, \bibinfo{person}{Jihai Chen}, \bibinfo{person}{Shuhua Yang}, \bibinfo{person}{Ali Al-Lawati}, \bibinfo{person}{Linli Tang}, \bibinfo{person}{Hui Liu}, {and} \bibinfo{person}{Suhang Wang}.} \bibinfo{year}{2025}\natexlab{b}.
\newblock \showarticletitle{A Survey on Collaborating Small and Large Language Models for Performance, Cost-effectiveness, Cloud-edge Privacy, and Trustworthiness}.
\newblock \bibinfo{journal}{\emph{arXiv}} (\bibinfo{year}{2025}).
\newblock
\href{https://doi.org/10.48550/arxiv.2510.13890}{doi:\nolinkurl{10.48550/arxiv.2510.13890}}


\bibitem[Wang et~al\mbox{.}(2023)]%
        {wang2023promptem}
\bibfield{author}{\bibinfo{person}{Pengwei Wang}, \bibinfo{person}{Zui Zeng}, {et~al\mbox{.}}} \bibinfo{year}{2023}\natexlab{}.
\newblock \showarticletitle{PromptEM: Prompt-tuning for Low-resource Generalized Entity Matching}.
\newblock \bibinfo{journal}{\emph{Proceedings of the VLDB Endowment}}  \bibinfo{volume}{16} (\bibinfo{year}{2023}).
\newblock


\bibitem[Wang et~al\mbox{.}(2025a)]%
        {DBLP:journals/dase/WangCWULG25}
\bibfield{author}{\bibinfo{person}{Xin Wang}, \bibinfo{person}{Zirui Chen}, \bibinfo{person}{Haofen Wang}, \bibinfo{person}{Leong~Hou U}, \bibinfo{person}{Zhao Li}, {and} \bibinfo{person}{Wenbin Guo}.} \bibinfo{year}{2025}\natexlab{a}.
\newblock \showarticletitle{Large Language Model Enhanced Knowledge Representation Learning: {A} Survey}.
\newblock \bibinfo{journal}{\emph{Data Sci. Eng.}} \bibinfo{volume}{10}, \bibinfo{number}{3} (\bibinfo{year}{2025}), \bibinfo{pages}{315--338}.
\newblock
\href{https://doi.org/10.1007/S41019-025-00285-Y}{doi:\nolinkurl{10.1007/S41019-025-00285-Y}}


\bibitem[Weng et~al\mbox{.}(2025)]%
        {weng2025bridgescope}
\bibfield{author}{\bibinfo{person}{Lianggui Weng} {et~al\mbox{.}}} \bibinfo{year}{2025}\natexlab{}.
\newblock \showarticletitle{BridgeScope: A Universal Toolkit for Bridging Large Language Models and Databases}.
\newblock \bibinfo{journal}{\emph{arXiv preprint arXiv:2508.04031}} (\bibinfo{year}{2025}).
\newblock


\bibitem[Wu et~al\mbox{.}(2023)]%
        {wu2023autogen}
\bibfield{author}{\bibinfo{person}{Qingyun Wu}, \bibinfo{person}{Gagan Bansal}, \bibinfo{person}{Jieyu Zhang}, \bibinfo{person}{Yiran Wu}, \bibinfo{person}{Shaokun Zhang}, \bibinfo{person}{Erkang Zhu}, \bibinfo{person}{Beibin Li}, \bibinfo{person}{Li Jiang}, \bibinfo{person}{Xiaoyun Zhang}, {and} \bibinfo{person}{Chi Wang}.} \bibinfo{year}{2023}\natexlab{}.
\newblock \showarticletitle{AutoGen: Enabling Next-Gen {LLM} Applications via Multi-Agent Conversation Framework}.
\newblock \bibinfo{journal}{\emph{CoRR}}  \bibinfo{volume}{abs/2308.08155} (\bibinfo{year}{2023}).
\newblock
\href{https://doi.org/10.48550/ARXIV.2308.08155}{doi:\nolinkurl{10.48550/ARXIV.2308.08155}}


\bibitem[Xiao et~al\mbox{.}(2024a)]%
        {xiao2023streamingllm}
\bibfield{author}{\bibinfo{person}{Guangxuan Xiao} {et~al\mbox{.}}} \bibinfo{year}{2024}\natexlab{a}.
\newblock \showarticletitle{Efficient Streaming Language Models with Attention Sinks}. In \bibinfo{booktitle}{\emph{The Twelfth International Conference on Learning Representations (ICLR)}}.
\newblock


\bibitem[Xiao et~al\mbox{.}(2024b)]%
        {xiao2023efficient}
\bibfield{author}{\bibinfo{person}{Guangxuan Xiao}, \bibinfo{person}{Yuandong Tian}, \bibinfo{person}{Beidi Chen}, \bibinfo{person}{Song Han}, {and} \bibinfo{person}{Mike Lewis}.} \bibinfo{year}{2024}\natexlab{b}.
\newblock \showarticletitle{Efficient Streaming Language Models with Attention Sinks}. In \bibinfo{booktitle}{\emph{The Twelfth International Conference on Learning Representations}}.
\newblock


\bibitem[Yang et~al\mbox{.}(2018)]%
        {yang2018hotpotqa}
\bibfield{author}{\bibinfo{person}{Zhilin Yang}, \bibinfo{person}{Peng Qi}, \bibinfo{person}{Saizheng Zhang}, \bibinfo{person}{Yoshua Bengio}, \bibinfo{person}{William~W. Cohen}, \bibinfo{person}{Ruslan Salakhutdinov}, {and} \bibinfo{person}{Christopher~D. Manning}.} \bibinfo{year}{2018}\natexlab{}.
\newblock \showarticletitle{{HotpotQA}: A Dataset for Diverse, Explainable Multi-Hop Question Answering}. In \bibinfo{booktitle}{\emph{Proceedings of the 2018 Conference on Empirical Methods in Natural Language Processing (EMNLP)}}. \bibinfo{pages}{2369--2380}.
\newblock
\urldef\tempurl%
\url{https://arxiv.org/abs/1809.09600}
\showURL{%
\tempurl}


\bibitem[Yao et~al\mbox{.}(2024)]%
        {yao2024cacheblend}
\bibfield{author}{\bibinfo{person}{Jiayi Yao}, \bibinfo{person}{Hanchen Li}, \bibinfo{person}{Yuhan Liu}, \bibinfo{person}{Siddhant Ray}, \bibinfo{person}{Yihua Cheng}, \bibinfo{person}{Qizheng Zhang}, \bibinfo{person}{Kuntai Du}, \bibinfo{person}{Shan Lu}, {and} \bibinfo{person}{Junchen Jiang}.} \bibinfo{year}{2024}\natexlab{}.
\newblock \showarticletitle{CacheBlend: Fast Large Language Model Serving for RAG with Cached Knowledge Fusion}. \bibinfo{publisher}{Association for Computing Machinery}, \bibinfo{address}{New York, NY, USA}.
\newblock
\href{https://doi.org/10.1145/3689031.3696098}{doi:\nolinkurl{10.1145/3689031.3696098}}


\bibitem[Yao et~al\mbox{.}(2025)]%
        {llmopt2025}
\bibfield{author}{\bibinfo{person}{Zhiming Yao}, \bibinfo{person}{Haoyang Li}, \bibinfo{person}{Jing Zhang}, \bibinfo{person}{Cuiping Li}, {and} \bibinfo{person}{Hong Chen}.} \bibinfo{year}{2025}\natexlab{}.
\newblock \showarticletitle{A Query Optimization Method Utilizing Large Language Models}.
\newblock \bibinfo{journal}{\emph{arXiv preprint arXiv:2503.06902}} (\bibinfo{year}{2025}).
\newblock


\bibitem[Ye et~al\mbox{.}(2025)]%
        {ye2025kvcomm}
\bibfield{author}{\bibinfo{person}{Hancheng Ye}, \bibinfo{person}{Zhengqi Gao}, \bibinfo{person}{Mingyuan Ma}, \bibinfo{person}{Qinsi Wang}, \bibinfo{person}{Yuzhe Fu}, \bibinfo{person}{Ming-Yu Chung}, \bibinfo{person}{Yueqian Lin}, \bibinfo{person}{Zhijian Liu}, \bibinfo{person}{Jianyi Zhang}, \bibinfo{person}{Danyang Zhuo}, {and} \bibinfo{person}{Yiran Chen}.} \bibinfo{year}{2025}\natexlab{}.
\newblock \showarticletitle{KVCOMM: Online Cross-context KV-cache Communication for Efficient LLM-based Multi-agent Systems}. In \bibinfo{booktitle}{\emph{Advances in Neural Information Processing Systems}}.
\newblock


\bibitem[Zhan et~al\mbox{.}(2025)]%
        {Zhanetal2025threemodel}
\bibfield{author}{\bibinfo{person}{Huiyou Zhan}, \bibinfo{person}{Xuan Zhang}, \bibinfo{person}{Haisheng Tan}, \bibinfo{person}{Han Tian}, \bibinfo{person}{Dongping Yong}, \bibinfo{person}{Junyang Zhang}, {and} \bibinfo{person}{Xiang-Yang Li}.} \bibinfo{year}{2025}\natexlab{}.
\newblock \showarticletitle{PICE: A Semantic-Driven Progressive Inference System for LLM Serving in Cloud-Edge Networks}.
\newblock \bibinfo{journal}{\emph{arXiv}} (\bibinfo{year}{2025}).
\newblock
\href{https://doi.org/10.48550/arxiv.2501.09367}{doi:\nolinkurl{10.48550/arxiv.2501.09367}}


\bibitem[Zhang et~al\mbox{.}(2024)]%
        {zhang2024jellyfish}
\bibfield{author}{\bibinfo{person}{Haochen Zhang}, \bibinfo{person}{Yuyang Dong}, \bibinfo{person}{Chuan Xiao}, {and} \bibinfo{person}{Masafumi Oyamada}.} \bibinfo{year}{2024}\natexlab{}.
\newblock \showarticletitle{Jellyfish: A Large Language Model for Data Quality}.
\newblock \bibinfo{journal}{\emph{Proceedings of the VLDB Endowment}} (\bibinfo{year}{2024}).
\newblock


\bibitem[Zhang et~al\mbox{.}(2025a)]%
        {zhang2025pqcache}
\bibfield{author}{\bibinfo{person}{Hailin Zhang}, \bibinfo{person}{Xiaodong Ji}, \bibinfo{person}{Yilin Chen}, \bibinfo{person}{Fangcheng Fu}, \bibinfo{person}{Xupeng Miao}, \bibinfo{person}{Xiaonan Nie}, \bibinfo{person}{Weipeng Chen}, {and} \bibinfo{person}{Bin Cui}.} \bibinfo{year}{2025}\natexlab{a}.
\newblock \showarticletitle{Pqcache: Product quantization-based kvcache for long context llm inference}.
\newblock \bibinfo{journal}{\emph{Proceedings of the ACM on Management of Data}} \bibinfo{volume}{3}, \bibinfo{number}{3} (\bibinfo{year}{2025}), \bibinfo{pages}{1--30}.
\newblock


\bibitem[Zhang et~al\mbox{.}(2025b)]%
        {zhang2025plforge}
\bibfield{author}{\bibinfo{person}{Hang Zhang}, \bibinfo{person}{Chaokun Wang}, \bibinfo{person}{Hongwei Li}, \bibinfo{person}{Cheng Wu}, \bibinfo{person}{Songyao Wang}, \bibinfo{person}{Yabin Liu}, \bibinfo{person}{Gengyuan Shi}, {and} \bibinfo{person}{Ziyang Liu}.} \bibinfo{year}{2025}\natexlab{b}.
\newblock \showarticletitle{PLForge: Enhancing Language Models for Natural Language to Procedural Extensions of SQL}.
\newblock \bibinfo{journal}{\emph{Proceedings of the ACM on Management of Data}} \bibinfo{volume}{3}, \bibinfo{number}{6} (\bibinfo{year}{2025}), \bibinfo{pages}{1--28}.
\newblock


\bibitem[Zhang et~al\mbox{.}(2026)]%
        {DBLP:conf/icde/ZhangLMWCLDC26}
\bibfield{author}{\bibinfo{person}{Xinyi Zhang}, \bibinfo{person}{Liangzu Liu}, \bibinfo{person}{Xupeng Miao}, \bibinfo{person}{Yinjun Wu}, \bibinfo{person}{Zhen Chen}, \bibinfo{person}{Wei Lu}, \bibinfo{person}{Xiaoyong Du}, {and} \bibinfo{person}{Bin Cui}.} \bibinfo{year}{2026}\natexlab{}.
\newblock \showarticletitle{DistVec: Efficient Distributed Machine Learning in Parallel Database Systems}. In \bibinfo{booktitle}{\emph{42nd {IEEE} International Conference on Data Engineering, {ICDE} 2026, Montreal, QC, Canada, May 4-8, 2026}}. \bibinfo{publisher}{{IEEE}}, \bibinfo{pages}{1--14}.
\newblock
\href{https://doi.org/10.1109/ICDE65706.2026.00169}{doi:\nolinkurl{10.1109/ICDE65706.2026.00169}}


\bibitem[Zhang et~al\mbox{.}(2023)]%
        {zhang2023h2o}
\bibfield{author}{\bibinfo{person}{Zhenyu Zhang} {et~al\mbox{.}}} \bibinfo{year}{2023}\natexlab{}.
\newblock \showarticletitle{H2O: Heavy-Hitter Oracle for Efficient Generative Inference of Large Language Models}. In \bibinfo{booktitle}{\emph{Advances in Neural Information Processing Systems (NeurIPS)}}.
\newblock


\bibitem[Zhao et~al\mbox{.}(2025)]%
        {zhao2025iam}
\bibfield{author}{\bibinfo{person}{Yi Zhao}, \bibinfo{person}{Zuchao Li}, {and} \bibinfo{person}{Hai Zhao}.} \bibinfo{year}{2025}\natexlab{}.
\newblock \showarticletitle{IAM: Efficient Inference through Attention Mapping between Different-scale LLMs}. In \bibinfo{booktitle}{\emph{Proceedings of the 63rd Annual Meeting of the Association for Computational Linguistics (Volume 1: Long Papers)}}.
\newblock


\bibitem[Zheng et~al\mbox{.}(2023)]%
        {zheng2023lmsyschat1m}
\bibfield{author}{\bibinfo{person}{Lianmin Zheng}, \bibinfo{person}{Wei-Lin Chiang}, \bibinfo{person}{Ying Sheng}, \bibinfo{person}{Tianle Li}, \bibinfo{person}{Siyuan Zhuang}, \bibinfo{person}{Zhanghao Wu}, \bibinfo{person}{Yonghao Zhuang}, \bibinfo{person}{Zhuohan Li}, \bibinfo{person}{Zi Lin}, \bibinfo{person}{Eric.~P Xing}, \bibinfo{person}{Joseph~E. Gonzalez}, \bibinfo{person}{Ion Stoica}, {and} \bibinfo{person}{Hao Zhang}.} \bibinfo{year}{2023}\natexlab{}.
\newblock \bibinfo{title}{LMSYS-Chat-1M: A Large-Scale Real-World LLM Conversation Dataset}.
\newblock
\showeprint[arxiv]{2309.11998}~[cs.CL]


\bibitem[Zheng et~al\mbox{.}(2024)]%
        {lian2024Sglang}
\bibfield{author}{\bibinfo{person}{Lianmin Zheng}, \bibinfo{person}{Liangsheng Yin}, \bibinfo{person}{Zhiqiang Xie}, \bibinfo{person}{Chuyue Sun}, \bibinfo{person}{Jeff Huang}, \bibinfo{person}{Cody~Hao Yu}, \bibinfo{person}{Shiyi Cao}, \bibinfo{person}{Christos Kozyrakis}, \bibinfo{person}{Ion Stoica}, \bibinfo{person}{Joseph~E. Gonzalez}, \bibinfo{person}{Clark~W. Barrett}, {and} \bibinfo{person}{Ying Sheng}.} \bibinfo{year}{2024}\natexlab{}.
\newblock \showarticletitle{SGLang: Efficient Execution of Structured Language Model Programs}. In \bibinfo{booktitle}{\emph{Advances in Neural Information Processing Systems 38: Annual Conference on Neural Information Processing Systems 2024, NeurIPS 2024, Vancouver, BC, Canada, December 10 - 15, 2024}}.
\newblock
\urldef\tempurl%
\url{http://papers.nips.cc/paper\_files/paper/2024/hash/724be4472168f31ba1c9ac630f15dec8-Abstract-Conference.html}
\showURL{%
\tempurl}


\bibitem[Zhou et~al\mbox{.}(2026)]%
        {zhou2026janus}
\bibfield{author}{\bibinfo{person}{Tianbao Zhou}, \bibinfo{person}{Yi Wang}, \bibinfo{person}{Yu Zhou}, \bibinfo{person}{Zirui Liu}, \bibinfo{person}{Zhiming Wang}, \bibinfo{person}{Yebo Peng}, \bibinfo{person}{Yongfu Wang}, \bibinfo{person}{Yi Zhang}, \bibinfo{person}{Jinrun Yin}, \bibinfo{person}{Kemeng Tian}, {et~al\mbox{.}}} \bibinfo{year}{2026}\natexlab{}.
\newblock \showarticletitle{Janus: Multi-LLM Serving at Production Scale}. In \bibinfo{booktitle}{\emph{Proceedings of the ACM SIGOPS 32nd Symposium on Operating Systems Principles}}. \bibinfo{pages}{1--17}.
\newblock


\bibitem[Zhu et~al\mbox{.}(2024)]%
        {zhu2024muxserve}
\bibfield{author}{\bibinfo{person}{Haojie Zhu} {et~al\mbox{.}}} \bibinfo{year}{2024}\natexlab{}.
\newblock \showarticletitle{MuxServe: Flexible Spatial-Temporal Multiplexing for Multiple LLM Serving}. In \bibinfo{booktitle}{\emph{41st International Conference on Machine Learning (ICML)}}.
\newblock


\bibitem[Zhu et~al\mbox{.}(2025)]%
        {DBLP:journals/dase/ZhuZSSY25}
\bibfield{author}{\bibinfo{person}{Jingyu Zhu}, \bibinfo{person}{Xintong Zhao}, \bibinfo{person}{Yu Sun}, \bibinfo{person}{Shaoxu Song}, {and} \bibinfo{person}{Xiaojie Yuan}.} \bibinfo{year}{2025}\natexlab{}.
\newblock \showarticletitle{Relational Data Cleaning Meets Artificial Intelligence: {A} Survey}.
\newblock \bibinfo{journal}{\emph{Data Sci. Eng.}} \bibinfo{volume}{10}, \bibinfo{number}{2} (\bibinfo{year}{2025}), \bibinfo{pages}{147--174}.
\newblock
\href{https://doi.org/10.1007/S41019-024-00266-7}{doi:\nolinkurl{10.1007/S41019-024-00266-7}}


\end{thebibliography}

\end{document}